\def\year{2022}\relax
\documentclass[letterpaper]{article} 
\usepackage{aaai22}  
\usepackage{times}  
\usepackage{helvet}  
\usepackage{courier}  
\usepackage[hyphens]{url}  
\usepackage{graphicx} 
\usepackage{natbib}  
\usepackage{caption} 
\DeclareCaptionStyle{ruled}{labelfont=normalfont,labelsep=colon,strut=off} 
\usepackage{soul}
\usepackage{color, colortbl}
\usepackage[utf8]{inputenc}
\usepackage{amsmath,amsfonts,amssymb}
\usepackage{booktabs}
\usepackage{subcaption}
\usepackage{latexsym}
\usepackage[noend]{algpseudocode}
\usepackage{xspace}
\usepackage{url}
\usepackage{enumerate}
\usepackage[defblank]{paralist}
\usepackage{microtype}
\usepackage{comment}
\usepackage{tabularx}
\usepackage{multirow}
\usepackage{MnSymbol}
\usepackage{stmaryrd}
\usepackage{cleveref}
\usepackage[linesnumbered,ruled,vlined]{algorithm2e}
\usepackage[switch]{lineno}
\usepackage{pgf}
\usepackage[mathscr]{euscript}
\usepackage{mathtools}
\usepackage{tikz}
\usepackage{soul}
\usepackage{makecell}

\usetikzlibrary{arrows, arrows.meta, patterns, automata}

\usepackage[inline]{enumitem}
\newlist{romanenumerate*}{enumerate*}{1}
\setlist[romanenumerate*]{label=(\textit{\roman*})}
\newlist{romanenumerate}{enumerate}{1}
\setlist[romanenumerate]{label=(\textit{\roman*})}

\usepackage{amsthm}
\usepackage{thmtools}
\usepackage{thm-restate}
\newtheorem{definition}{Definition}

\newtheorem{theorem}{Theorem}

\SetKwFunction{KwFn}{Fn}
\SetKwInOut{Input}{Input}
\SetKwInOut{Output}{Output}
\SetKwComment{Comment}{\tiny \#~}{}

\definecolor{grayline}{RGB}{244,246,246}

\usepackage{todonotes}

\newcommand{\BLIND}{\mathit{h}^{\textsc{blind}}\xspace}
\newcommand{\hADD}{\mathit{h}^{\textsc{add}}\xspace}
\newcommand{\hFF}{\mathit{h}^{\textsc{ff}}\xspace}
\newcommand{\hMAX}{\mathit{h}^{\textsc{max}}\xspace}

\newcommand{\unsolvable}{\textsc{unsolvable}}
\newcommand{\solved}{\textsc{solved}}
\newcommand{\unsolved}{\textsc{unsolved}}

\newcommand{\reachedfixedpoint}{\textsc{reachedFixedPoint}}
\newcommand{\promising}{\textsc{promising}}

\newcommand{\true}{\textsc{true}}
\newcommand{\false}{\textsc{false}}

\newcommand{\flag}{\mathit{flag}}

\newcommand{\bound}{\mathit{bound}}
\newcommand{\newbound}{\mathit{nextBound}}

\newcommand{\ancestorsset}{\mathscr{Z}}

\newcommand{\successorsset}{\mathscr{M}}
\newcommand{\solvedsuccessorsset}{\mathscr{M}_*}

\newcommand{\nonpromisingset}{\mathscr{X}}
\newcommand{\applicableactions}{\textsc{ApplicableActions}}

\newcommand{\maxF}{\F_{\max}\xspace}
\newcommand{\minF}{\F_{\min}\xspace}

\newcommand{\emptypartialfunction}{\emptyset}
\newcommand{\emptypolicy}{\emptypartialfunction}

\newcommand{\s}{\lbrace s \rbrace}
\newcommand{\ancestorswiths}{\ancestorsset \cup \s}

\newcommand{\LAOStar}{LAO$^{*}$}

\newcommand{\IDFSSatisficing}{{\sc idfsp}\xspace}
\newcommand{\IDFS}{{\sc idfs}\xspace}

\newcommand{\IDFStitle}{\textbf{\textsc{idfs}}\xspace}
\newcommand{\IDFSRectitle}{\textbf{\textsc{idfs$_\mathbf{R}$}}\xspace}

\newcommand{\IDFSP}{{\sc idfsp}\xspace}

\newcommand{\IDFSRec}{{\sc idfs$_R$}\xspace}
\newcommand{\IDFSPRec}{{\sc idfsp$_R$}\xspace}

\newcommand{\variables}{\V}
\newcommand{\initialstate}{s_0}
\newcommand{\goalcondition}{s_*}
\newcommand{\actions}{\A}

\newcommand{\pre}{\emph{pre}}
\newcommand{\eff}{\mathit{eff}}
\newcommand{\effs}{\textsc{EFFS}}
\newcommand{\suc}{\textsc{succ}}
\newcommand{\succs}{\textsc{SUCCS}}

\newcommand{\cv}{\mathbf{cv}}

\newcommand{\FONDSAT}{{\sc fondsat}\xspace}

\newcommand{\PRP}{{\sc prp}\xspace}

\newcommand{\Paladinus}{{\sc paladinus}\xspace}

\newcommand{\MYND}{{\sc myND}\xspace}

\newcommand{\GRENDEL}{{\sc grendel}\xspace}

\newcommand{\FIP}{{\sc fip}\xspace}

\newcommand{\GAMER}{{\sc gamer}\xspace}

\newcommand{\MBP}{{\sc mbp}\xspace}

\newcommand{\NDP}{{\sc ndp}\xspace}

\newcommand{\acrobatics}{{\sc acrobatics}\xspace}
\newcommand{\beamwalk}{{\sc beam-walk}\xspace}
\newcommand{\bwtwo}{{\sc bw-2}\xspace}
\newcommand{\bworiginal}{{\sc bw-orig}\xspace}
\newcommand{\bwnew}{{\sc bw-new}\xspace}
\newcommand{\chain}{{\sc chain}\xspace}
\newcommand{\doors}{{\sc doors}\xspace}
\newcommand{\earthobs}{{\sc earth-obs}\xspace}
\newcommand{\elevators}{{\sc elevators}\xspace}
\newcommand{\faults}{{\sc faults}\xspace}

\newcommand{\firstresp}{{\sc first-resp}\xspace}
\newcommand{\islands}{{\sc islands}\xspace}
\newcommand{\miner}{{\sc miner}\xspace}
\newcommand{\twspiky}{{\sc tw-spiky}\xspace}
\newcommand{\twtruck}{{\sc tw-truck}\xspace}
\newcommand{\tritw}{{\sc tri-tw}\xspace}
\newcommand{\zeno}{{\sc zeno}\xspace}

\newcommand{\A}{\mathcal{A}} 
 
 \newcommand{\F}{\mathcal{F}}

 \renewcommand{\P}{\mathcal{P}}
 
\renewcommand{\S}{\mathcal{S}} 
 \newcommand{\V}{\mathcal{V}}

\newcommand{\commentout}[1]{}

\newcommand{\FOND}{{\sc fond}\xspace}

\newcommand{\FONDtitle}{\textbf{\textsc{fond}}\xspace}

\newcommand{\Nat}{{\rm I\kern-.23em N}}

\newcolumntype{X}{>{\centering}p{0.8mm}}

\title{Iterative Depth-First Search for Fully Observable Non-Deterministic Planning}
\author{
Ramon Fraga Pereira$^{1}$, André Grahl Pereira$^{2}$, Frederico Messa$^2$, Giuseppe De Giacomo$^1$\\
}
\affiliations{
$^1$Sapienza University of Rome, Rome, Italy\\
$^2$Federal University of Rio Grande do Sul, Porto Alegre, Brazil\\
$\lbrace$pereira,degiacomo$\rbrace$@diag.uniroma1.it \\ $\lbrace$agpereira,frederico.messa$\rbrace$@inf.ufrgs.br
}
\begin{document}

\maketitle

\begin{abstract}

\emph{Fully Observable Non-Deterministic} (\FOND) planning models uncertainty through actions with \emph{non-deterministic} effects.
Existing \FOND planning algorithms are effective and employ a wide range of techniques. However, most of the existing algorithms are not robust for dealing with both non-determinism and task size.
In this paper, we develop a novel \emph{iterative depth-first search} algorithm that solves \FOND planning tasks and produces \emph{strong cyclic policies}. Our algorithm is explicitly designed for \FOND planning, addressing more directly the non-deterministic aspect of \FOND planning, and it also exploits the benefits of heuristic functions to make the algorithm more effective during the iterative searching process.
We compare our proposed algorithm to well-known \FOND planners, and show that it has robust performance over several distinct types of \FOND domains considering different metrics.

\end{abstract}

\section{Introduction}\label{sec:introduction}

\emph{Fully Observable Non-Deterministic} (\FOND) planning is an important planning model that aims to handle the uncertainty of the effects of actions~\cite{CimattiPRT03}. In \FOND planning, states are \emph{fully observable} and actions may have \emph{non-deterministic} effects (i.e., an action may generate a set of possible successor states). \FOND planning is relevant for solving other related planning models, such as \emph{stochastic shortest path} (SSP) planning~\cite{BertsekasT91}, \emph{planning for temporally extended goals}~\cite{PatriziLG_IJCAI13,CTMBM17,CamachoM19_IJCAI,CamachoBMM18,ijcai2018GiuseppeSasha,brafman2019planning}, and \emph{generalized planning}~\cite{GeneralizedP_HuG11,GenerelizedIjcai2017BonetEtAl,kr2020bdgp}. 
Solutions for \FOND planning can be characterized as \emph{strong policies} which guarantee to achieve the goal condition in a finite number of steps, and \emph{strong cyclic policies}
which guarantee to lead only to states from
which a goal condition is satisfiable in a finite number of steps~\cite{CimattiPRT03}.

Existing \FOND planning algorithms in the literature are based on a diverse set of techniques and effectively solve difficult tasks when the non-determinism of the actions must be addressed. \citet{CimattiPRT03} and \citet{Gamer_KissmannE09} have introduced \emph{model-checking} planners based on \emph{binary decision diagrams}. Some of the most effective \FOND planners rely on standard \emph{Classical Planning} techniques by enumerating plans for a deterministic version of the task until producing a strong cyclic  policy~\cite{NDP_KuterNRG08,FIP_FuNBY11,Muise12ICAPSFond,MuiseICAPS14,MuiseAAAI14}.
There are also planners that efficiently employ AND/OR heuristic search for solving \FOND planning tasks, such as \MYND~\cite{MyND_MattmullerOHB10} and \GRENDEL~\cite{Grendel_RamirezS14}.
Recently, \citet{GeffnerG18_FONDSAT} have proposed a SAT encoding for \FOND planning, and an iterative SAT-based planner that effectively handles the uncertainty of \FOND planning. Nevertheless, these \FOND planners present some limitations. Some of these planners address the non-determinism of the actions more indirectly, whereas others rely on algorithms with sophisticated and costly control procedures, and others do not take advantage of fundamental characteristics of planning models. As a result, such \FOND planners are not robust for dealing with some of the non-determinism aspects of \FOND planning and task size.

In this paper, we introduce a novel \emph{iterative depth-first search} algorithm that solves \FOND planning tasks and produces \emph{strong cyclic policies}.
Our algorithm is based on two main concepts: (1) it is explicitly designed for solving \FOND planning tasks, so it addresses more directly the non-deterministic aspect of \FOND planning during the searching process; and (2) it exploits the benefits of heuristic functions to make the iterative searching process more effective.
We also introduce an efficient version of our algorithm that prunes unpromising states in each iteration.
To better understand the behavior of our proposed \emph{iterative depth-first search} algorithm, we characterize its behavior through fundamental properties of \FOND planning policies.

We empirically evaluate our algorithm over two \FOND benchmark sets: a set from IPC~\cite{IPC6}~and~\cite{Muise12ICAPSFond}; and a set containing new \FOND planning domains, proposed by \citet{GeffnerG18_FONDSAT}.
We show that our algorithm outperforms some of the existing state-of-the-art \FOND planners on planning time and coverage, especially for the new \FOND domains.
We also show that the pruning technique makes our algorithm competitive with existing \FOND planners.
Our contributions open new research directions in \FOND planning, such as the design of more informed heuristic functions, and the development of more effective search algorithms.

\section{Background}\label{sec:background}

\subsection{\FONDtitle Planning}

A \emph{Fully Observable Non-Deterministic} (\FOND) planning task~\cite{MyND_MattmullerOHB10} is a tuple $\Pi = \langle \variables, \initialstate, \goalcondition, \actions \rangle$. $\variables$ is a set of \emph{state variables}, and each variable $v \in \variables$ has a finite domain $\textsc{D}_{v}$. A \emph{partial state}~$s$ maps variables $v\in\V$ to values in~$\textsc{D}_{v}$, $s[v] \in \textsc{D}_{v}$,  or to a undefined value~$s[v] = \perp$.  $\mathit{vars}(s)$ is the set of variables in~$s$ with defined values.
If every variable~$\V$ in~$s$ is defined, then~$s$ is a \emph{state}.  $\initialstate$ is a state representing the \emph{initial} state, whereas $\goalcondition$ is a partial state representing the \emph{goal condition}. A state~$s$ is a \emph{goal} state if and only if $s \models s_*$.
$\actions$ is a finite set  of \emph{non-deterministic actions}, in which every action $a \in \A$ consists of $a = \langle \pre, \effs \rangle$, where $\pre(a)$ is a partial state called \emph{preconditions}, and $\effs(a)$ is a non-empty 
set of partial states that represent the possible \emph{effects} of $a$.
A \emph{non-deterministic action} $a \in \A$ is applicable in a state $s$ iff $s \models \pre(a)$. The \emph{application} of an effect $\eff \in \effs(a)$ to a state~$s$ generates a state $s' = \suc(s, \eff)$ with $s'[v] = \eff[v]$ if $v \in \mathit{vars}(\eff)$, and $s'[v]=s[v]$ if not.
The application of $\effs(a)$ to a state $s$ generates a set of successor states $\succs(s, a) = \lbrace \suc(s, \eff)~\mid \eff \in \effs(a) \rbrace$. We call $a\in \A$ simple \emph{deterministic} if $|\effs(a)|$ has size one.

A solution to a \FOND planning task $\Pi$ is a \emph{policy}~$\pi$ which is formally defined as a partial function $\pi: \S \mapsto \A \cup \lbrace \perp \rbrace$, which maps non-goal states of $\S$ into actions, such that an action $\pi(s)$ is applicable in the state $s$.
A $\pi$-trajectory with length $k-1$ is a non-empty sequence of states $\langle s^1, s^2, \dots s^k \rangle$, such that $s^{i+1} \in \succs(s^i, \pi(s^i)), \forall i \in \lbrace 1, 2, \dots, k-1 \rbrace$. A $\pi$-trajectory is called empty if it has a single state, and thus length zero.
A policy~$\pi$ is \emph{closed} if any $\pi$-trajectory starting from $s_0$ ends either in a goal state or in a state defined in the policy~$\pi$.
A policy~$\pi$ is a \emph{strong policy} for $\Pi$ if it is closed and no $\pi$-trajectory passes through a state more than once.
A policy $\pi$ is a \emph{strong cyclic policy} for $\Pi$ if it is closed and any $\pi$-trajectory starting from $s_0$ which does not end in a goal state, ends in a state $s'$ such that exists another $\pi$-trajectory starting from $s'$ ending in a goal state.
Note that a \emph{strong cyclic policy} may re-visit states infinite times, in a cyclic way, but the \emph{fairness} assumption guarantees that it will \emph{almost surely} reach a goal state at some point along the execution. The assumption of \emph{fairness} defines that all action outcomes in a given state will occur infinitely often~\cite{CimattiPRT03}.

\subsection{Determinization and Heuristics for \FONDtitle Planning}

A \emph{determinization} of a \FOND planning task $\Pi$ defines a new \FOND planning task~$\Pi^{\textsc{Det}}$ where all actions are \emph{deterministic}. Formally, $\Pi^{\textsc{Det}} = \langle \variables, \initialstate, \goalcondition, \actions^{\textsc{Det}} \rangle$ is a task where $\actions^{\textsc{Det}}$ is a set of deterministic actions with one action~$a'$ for each outcome~$\eff\in\effs(a)$ of all actions in $a\in\actions$. A $s$-plan for~$\Pi^{\textsc{Det}}$ is a sequence of actions that when applied to~$s$ reaches a goal state. A $s$-plan is optimal if it has minimum cost among all $s$-plans. A solution for~$\Pi^{\textsc{Det}}$ is a $\initialstate$-plan.

A \emph{heuristic function} $h: \S \mapsto \mathbb{R} \cup \{\infty\}$ maps a state~$s$ to its $h$-value, an estimation of the cost of a $s$-plan. A \emph{perfect heuristic}~$h^*$ maps a state~$s$ to its \textit{optimal} cost plan or $\infty$, if no plan exists. A heuristic is \emph{admissible} if $h(s) \leq h^*(s)$ for all $s \in \S$. \emph{Delete-relaxation} heuristics~\cite{Bonet01planning_AIJ,HoffmannN01_JAIR} can be efficiently used in \FOND planning by applying \emph{determinization}~\cite{Mattmueller_thesis_2013}.
Other types of heuristics for \FOND planning have been proposed in the literature, such as \emph{pattern-database} heuristics~\cite{MyND_MattmullerOHB10}, and pruning techniques~\cite{FOND_H_WintererW016,FOND_H_WintererA0W17}.

\subsection{\FONDtitle Planners}

One of the first \FOND planners in the literature was developed by~\citeauthor{CimattiPRT03}~\shortcite{CimattiPRT03}, and it is called \MBP (Model-Based Planner). \MBP solves \FOND planning tasks via \emph{model-checking}, and it is built upon \emph{binary decision diagrams} (BDDs). \GAMER~\cite{Gamer_KissmannE09}, the winner of the \FOND track at IPC~\shortcite{IPC6}, is also based on BDDs, but \GAMER has shown to be much more efficient than \MBP.

\MYND~\cite{MyND_MattmullerOHB10} is a \FOND planner based on an adapted version of \LAOStar~\cite{LAOStarHansenZ01a}, a heuristic search algorithm that has theoretical guarantees to extract strong cyclic solutions for Markov decision problems.
\NDP~\cite{NDP_KuterNRG08} makes use of \emph{Classical Planning} algorithms to solve \FOND planning tasks.
\FIP \cite{FIP_FuNBY11} is similar to \NDP, but the main difference is that \FIP avoids exploring already explored/solved states, being more efficient than \NDP.
\PRP~\cite{Muise12ICAPSFond} is one the most efficient \FOND planners in the literature, and it is built upon some improvements over the state relevance techniques, such as avoiding dead-ends states.
The main idea of these planners is selecting a reachable state~$s$ by the current policy that still is undefined in the current policy. Then, the planner finds a $s$-plan with $\Pi^{\textsc{Det}}$ and incorporates the $s$-plan into the policy. The planner repeats this process until the policy is strong cyclic, or it finds out that it is not possible to produce a strong cyclic policy from the current policy, and then it backtracks. Since these planners find $s$-plan for $\Pi^{\textsc{Det}}$ which do not consider the non-deterministic effects, they can take too much time to find that the current policy can not become a strong cyclic policy, or they can add actions to a policy that require too much search effort to become a strong cyclic policy.

\GRENDEL \cite{Grendel_RamirezS14} is a \FOND planner that combines regression with a symbolic fixed-point computation for extracting strong cyclic policies.
Most recently, \citet{GeffnerG18_FONDSAT} developed \FONDSAT, an iterative SAT-based \FOND planner that is capable to produce strong and strong cyclic policies for \FOND planning tasks.

\section{Iterative Depth-First Search Algorithm \\ for \FONDtitle Planning}\label{sec:idfs-search}

In this section, we propose a novel \textit{iterative depth-first search}
algorithm called \IDFS that produces strong cyclic policies
for \FOND planning tasks. \IDFS performs a series of bounded
depth-first searches that consider the non-determinism aspect
of \FOND planning during the iterative searching process. \IDFS
produces a strong cyclic policy in a bottom-up way and only
adds an action to the policy if it determines that the resulting
policy with the additional action has the potential to become
a strong cyclic policy without exceeding the current search-
depth bound.

\subsection{Evaluation Function~$\F$}

A heuristic function $h(s)$ estimates the length of a trajectory from the state~$s$ to any goal state. It can assess whether a search procedure can reach a goal state without exceeding a search-depth bound. We define the $f$-value of a state~$s$ as $f(s) = g(s) + h(s)$, with $g(s)$ being the search depth from $\initialstate$ to~$s$. In this paper, we assume that all actions have a uniform action cost equal to one\footnote{All \FOND planning domains in the available benchmarks have actions with unitary cost.}.
During the iterative searching process, \IDFS considers the application of an action $a \in \A$ to a state~$s$ by evaluating the set of generated successor states~$\succs(s,a)$ using an \emph{evaluation function $\F_\xi$}, which returns the estimate of the search depth required to reach a goal state through $\succs(s,a)$. The evaluation function $\F_\xi$ uses a parameter function~$\xi$ to aggregate the $f$-values of states in $\succs(s,a)$: $\F_{\min}(\succs(s,a))$ is $\min_{s' \in \succs(s,a)} f(s')$, and $\F_{\max}(\succs(s,a))$ is $\max_{s' \in \succs(s,a)} f(s')$.  Note that the evaluation function $\F_\xi$ is ``pessimistic'' when $\xi = \max$, whereas it is ``optmistic'' when $\xi = \min$.

\subsection{The \IDFStitle Algorithm}

We now present the \IDFS, and Algorithm~\ref{alg:IDFS} formally shows its pseudo-code.


\SetKwFunction{IDFSS}{\sc idfs}
\SetKwFunction{IDFSSearch}{\sc idfs}
\SetKwFunction{IDFSSearchRec}{\sc idfs$_R$}
\SetKwProg{Fn}{}{:}{}

\begin{algorithm}[t!]
\SetInd{0.27em}{0.54em}
\fontsize{8.5}{9.5}\selectfont
\caption{\IDFS}\label{alg:IDFS}
\DontPrintSemicolon
\tcp{\color{darkgray} Main Iterative Loop.}
\Fn{\IDFSSearch{$s_0$}}
{
    $\bound \coloneqq h(s_0), \newbound \coloneqq \infty$\;
    \While{$\bound \leq |\S|$}
    {
        $\flag, \pi \coloneqq \IDFSSearchRec(s_0, \emptyset, \emptyset, \emptypolicy)$\;
        \If{$\flag = \solved$}
        {
            \Return{$\pi$}
        }
        $\bound \coloneqq \newbound, \newbound \coloneqq \infty$
    }
    \Return{$\unsolvable$}
}
\tcp{\color{darkgray} Recursion.}
\Fn{\IDFSSearchRec{$s, \ancestorsset, \ancestorsset_*, \pi$}}
{
    \tcp{\color{darkgray} Base Cases.}
    \If{$s \models \goalcondition$ \textnormal{\bf or} $\pi(s) \neq \bot$ \textnormal{\bf or} $s \in \ancestorsset_*$}
    {
        \Return{$\solved, \pi$}]\label{alg:IDFS:solved_1}
    }
    \If{$s \in (\ancestorsset \setminus \ancestorsset_*)$}
    {
        \Return{$\unsolved, \pi$}
    }
    \tcp{\color{darkgray} Evaluate Actions.}
    \For{$a \in \applicableactions(s)$}
    {
        \If{$\F_\xi(\succs(s, a)) > \bound$ \textnormal{\bf and} $\ancestorsset_* = \emptyset$}
        {
            $\newbound \coloneqq \min \lbrace \newbound, \F_\xi(\succs(s, a)) \rbrace$\;
            \bf{continue} \tcp{\color{darkgray} Next action.}
        }
        \If{$g(s)+1 > \bound$}
        {
            $\newbound \coloneqq \min \lbrace \newbound, g(s)+1 \rbrace$\;
            \bf{continue} \tcp{\color{darkgray} Next action.}
        }
        \tcp{\color{darkgray} Fixed Point.}
        $\ancestorsset_*' \coloneqq \ancestorsset_*, \pi' \coloneqq \pi, \successorsset \coloneqq \succs(s,a), \solvedsuccessorsset \coloneqq \emptyset$\;
        \Repeat{$\reachedfixedpoint$}
        {
            $\reachedfixedpoint \coloneqq \true$\;
            \For{$s' \in (\successorsset \setminus \solvedsuccessorsset) $}
            {
                $\flag, \pi' \coloneqq \IDFSSearchRec(s', \ancestorsset \cup \lbrace s \rbrace, \ancestorsset_*', \pi')$\;
                \If{$\flag = \solved$}
                {
                    $\solvedsuccessorsset \coloneqq \solvedsuccessorsset \cup \lbrace s' \rbrace$\;
                    $\ancestorsset_*' \coloneqq \ancestorsset \cup \lbrace s \rbrace$\;
                    $\reachedfixedpoint \coloneqq \false$\;
                }
            }
        }
        \If{$\solvedsuccessorsset = \successorsset$}
        {
            $\pi'(s) \coloneqq a$\;
            \Return{$\solved, \pi'$}\label{alg:IDFS:solved_2}
        }
    }
    \Return{$\unsolved, \pi$}
}
\end{algorithm}


\subsubsection{Main Iterative Loop (Lines 1-8)}

\IDFS performs a series of bounded depth-first searches, called \emph{iterations} to solve a \FOND planning task $\Pi$.
\IDFS assumes that~$h$ is a heuristic function for the \emph{deterministic} version of the task~$\Pi$.
Prior to the first iteration, \IDFS initializes the \emph{bound} with the estimated value of heuristic function~$h$ of the initial state~$\initialstate$.
At each iteration, \IDFS aims to produce a solution by searching to a depth of at most \emph{bound}.
The main loop receives a flag indicating if the iteration produced a solution from state~$s_0$.
If the flag is $\solved$, then~$\pi$ is a strong cyclic policy for task~$\Pi$, and \IDFS returns it.
If the flag is $\unsolved$,  then \IDFSRec could not produce a strong cyclic policy for task~$\Pi$ with the current \emph{bound}. Thus, \IDFS assigns to~$\bound$ the value of the global variable~$\newbound$.
The value of $\newbound$ is the minimum estimate ($\F_\xi$ or $g$-value$+1$) of a generated but not expanded set of successors.
If no set of successors with a greater estimate than~$\bound$ is generated, the main loop returns $\unsolvable$. This general strategy of depth-first search bounded by estimates is inspired by the Iterative Deepening~A$^*$ algorithm by~\citet{IDAStar_Korf85}.

\subsubsection{Recursion (Lines 9-34)}

\IDFS iteratively tries to produce a strong cyclic policy for task $\Pi$ in a bottom-up way, using a recursive procedure called \IDFSRec. Definition~\ref{def:partial_cyclic_policy} formally defines the concept of \emph{partial strong cyclic policy}, which we use to explain the behavior of \IDFS.

\begin{definition}\label{def:partial_cyclic_policy}
    A policy~$\pi$ is a \textbf{partial strong cyclic policy} from a state~$s$ of a \FOND task~$\Pi$ for a set~$A$ of primary target states and a set~$B$ of secondary target states, iff $A$ is~\textbf{reachable} from~$s$ in $\pi$, and $\pi$ is \textbf{sinking} to $B$. (We omit $A$ and $B$, when the context is clear.)

    \begin{itemize}
        \item $A$ is \textbf{reachable} from~$s$ in $\pi$ iff $s \in A$ or there is a $\pi$-trajectory starting from~$s$ ending in a state of $A$ that does not includes a state of $B\setminus A$.
        \item $\pi$ is \textbf{sinking} to $B$ iff any $\pi$-trajectory either goes through a state of $B$ or ends in a state $s'$, such that exists another $\pi$-trajectory starting from $s'$ ending in a state of $B$.
    \end{itemize}
\end{definition}

\IDFSRec aims to produce a \emph{partial strong cyclic policy} from state~$s$ of a \FOND task~$\Pi$ by searching to a depth of at most the current~$\bound$. \IDFSRec takes as input four arguments: the state~$s$, the set $\ancestorsset$, the set $\ancestorsset_*$, and a policy $\pi$. These arguments are set to empty in each iteration of the main iterative loop.
The set $\ancestorsset$ contains the ancestors of state~$s$.
The policy~$\pi$ is the policy that \IDFS has built up to the moment of the current call of \IDFSRec. The set $\ancestorsset_*\subseteq\ancestorsset$ contains all ancestors of state~$s$ that:~are not in $\pi$ and are ancestors of states in~$\pi$. Note that, in this case, \IDFS has found a trajectory to a goal state for all states in~$\ancestorsset_*$.
If \IDFSRec returns~$\solved$, then the returned policy is a \emph{partial strong cyclic policy} from state~$s$. The sets $A$ and $B$ of the partial strong cyclic policy are $A = \S_* \cup \mathcal{S}_\pi \cup \ancestorsset_*$ (primary target states) and the set $B = A \cup \ancestorsset$ (secondary target states). $\mathcal{S}_\pi$ is the set $\lbrace s \mid \pi(s) \neq \bot \rbrace$, and $\S_*$ the set $\lbrace s \mid s \models \goalcondition \rbrace$ of goal states. Since $A = B = \S_*$ in the call of \IDFSRec in the main loop, the returned partial strong cyclic policy from state~$s_0$ is a \emph{strong cyclic policy} for task~$\Pi$.

Consider the \FOND planning task example of Figure~\ref{fig:algorithm_example}, in which, $s_0$ is the initial state, $s_5$ is the only goal state, and there are three non-deterministic actions, applied in states~$s_1, s_3$ and $s_{10}$.
In the example, the current call of \IDFSRec is evaluating the state~$s_{10}$ (with the current recursion path is in bold). In this call, the received policy~$\pi$ contains the states $s_2$, $s_3$, $s_4$, $s_6$ and $s_8$ (in purple), i.e., $\S_\pi=\{s_2, s_3, s_4, s_6, s_8\}$. The ancestors of state~$s_{10}$ are the states $s_0$, $s_1$ and $s_9$, i.e., $\ancestorsset=\{s_0, s_1, s_9\}$. The former two are in $\ancestorsset_*=\{s_0, s_1\}$ (in green). Thus, the set~$B$ of secondary target states includes states of~$A$ and the state~$s_9$.

\subsubsection{\IDFSRectitle Base Cases (Lines 10-13)}

\IDFSRec first checks whether the current state~$s$ of the recursion is either a primary target state ($s \models \goalcondition$ or $s \in \ancestorsset_*$ or $\pi(s) \neq \bot$), or a state~$s\in \ancestorsset$ which is not primary target state. If the first case occurs, \IDFSRec returns $\solved$. If the second case occurs, it returns $\unsolved$. Both cases return policy unmodified.

\subsubsection{\IDFSRectitle Evaluate Actions (Lines 14-20)}

If the base cases do not address the state~$s$, \IDFSRec proceeds to attempt to solve it (Line~14).
To optimize the search, \IDFSRec evaluates first the \emph{applicable actions} with least $\F_{\max}(\succs(s,a))$ and discards actions with $\F_{\max}(\succs(s,a))=\infty$.
If the estimated solution depth of the successor states~$\succs(s,a)$ is greater than the current $\bound$ (i.e., $\F_\xi(\succs(s,a))> \bound$) and $\ancestorsset_* = \emptyset$ (Line~15), then the set of successor states is discarded, and $\F_\xi(\succs(s,a))$ is assigned to $\newbound$ (Line~16) if $\newbound$ was greater than it.

\IDFSRec verifies whether $\ancestorsset_* = \emptyset$ because it aims to find at least one trajectory from~$s$ to a primary target state in $A = \S_* \cup \mathcal{S}_\pi \cup \ancestorsset_*$. Note that $\F_\xi(\succs(s,a))$ aggregates $f$-values that only estimate the solution depth from $s_0$ through $\succs(s,a)$ to goal states. Thus, $\F_\xi(\succs(s,a))$ can only be used to estimate the solution depth to a primary target state when $\ancestorsset_* = \emptyset$, since it implies $A = \S_*$.

If $\ancestorsset_* \neq \emptyset$  the $g$-value the successor states~$\succs(s,a)$ can be used to estimate the solution depth to a primary target state.
In this case, if $g(n) + 1$ is greater than the current $\bound$, the set of successor states is discarded, and $g(n) + 1$ is assigned to $\newbound$ if $\newbound$ was greater than it.
If neither $\F_\xi(\succs(s,a))$ nor $g(s)$ prevent the search to proceed, \IDFSRec evaluates the successor states $\succs(s,a)$.

\subsubsection{\IDFSRectitle Fixed Point (Lines 21-33)}

\begin{figure}[t]
    \centering
    \includegraphics[width=.8\columnwidth]{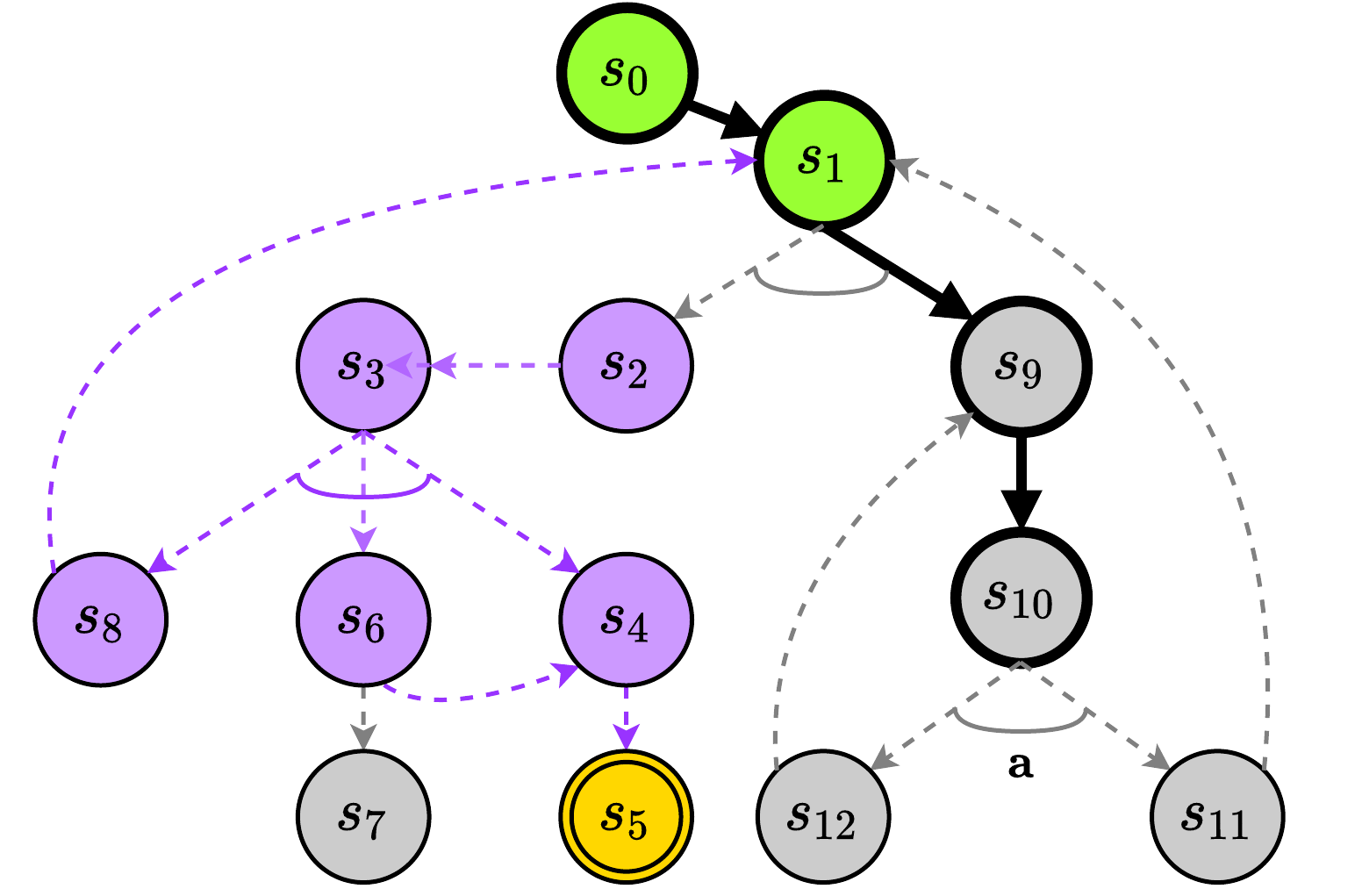}
	\caption{Current path of analysis is in bold. Goal state in yellow. $\pi$ is in purple. $\ancestorsset_*$ is in green.}
	\label{fig:algorithm_example}
\end{figure}

\IDFSRec recursively descends into the successor states $\succs(s,a)$ of $s$ to determine whether it should or not add the mapping $s \mapsto a$ to the police $\pi$. Namely, it adds the mapping $s \mapsto a$ to $\pi$ only if \textbf{all} the recursive calls on states of $\succs(s,a)$ returned $\solved$ (Lines~31--32). If not, it discards the possibility of using the action~$a$ on $s$, and proceeds to the next action.

Consider again the \FOND planning task example of Figure~\ref{fig:algorithm_example}. Assume that \IDFSRec reaches the point to evaluate the successor states~$s_{11}$ and $s_{12}$ of $s_{10}$.
Note that for~$s_{10}$ the set of~$A$ primary target states is $\{s_0, s_1,s_2, s_3, s_4, s_5, s_6, s_8\}$, and the set~$B$ of secondary target states is $\{s_0, s_1,s_2, s_3, s_4, s_5, s_6, s_8,s_9\}$.
\IDFSRec aims to produce a policy~$\pi'$ such that~$A$ is reachable from $s_{10}$ in $\pi'$, and $\pi'$ is \emph{sinking} to~$B$. To ensure that, \IDFSRec must find a trajectory from $s_{10}$ to a state in~$A$ that does not include a state of $B \setminus A$.
Thus, \IDFSRec analyzes all successors of~$s_{10}$ to find such a trajectory.
Before finding this trajectory, the arguments $\pi' \coloneqq \pi$, $\ancestorsset' \coloneqq \ancestorswiths$ and $\ancestorsset_*' \coloneqq \ancestorsset_*$ passed to \IDFSRec when evaluating states~$s_{11}$ and $s_{12}$ remaining unchanged.

Suppose the first recursive call evaluates $s_{12}$, thus the primary targets states for $s_{12}$ are $A$. Since $s_9$ is an ancestor of $s_{12}$ and $s_9 \notin A$, there is no trajectory from $s_{12}$ to a state of $A$ that does not includes a state of $B\setminus A$. Thus, the recursive call will fail and return $\unsolved$.
Next, \IDFSRec will proceed to the other successor state of $s_{10}$, namely $s_{11}$.
If the recursive call on $s_{11}$ fails because of the bound, the algorithm will have analyzed all successors of $s_{10}$ without having any progress, as the set of successors states ``already solved'' $\solvedsuccessorsset$ would not have changed, and thus a fixed-point would be reached, resulting in the action being discarded.

Suppose the recursive call on state~$s_{11}$ does not fail, and it returns $\solved$. Then, the returned policy is a partial strong cyclic policy from $s_{11}$ for the set of primary states $A'$ and the set of secondary states $B'$. Since $A' = A$ and $B' = B \cup \s$.
\IDFSRec now evaluates $s_{12}$ again, but now with a modified $A'$. Since we already have a trajectory from $s_{10}$ to $A$, now $A'$  includes also $s_9, s_{10}$ and $s_{11}$, and  $B'=A'$. The recursive call on $s_{12}$ returns $\solved$ because there is a trajectory to $A'$. The new policy extended with $s_{10} \mapsto \mathbf{a}$ is a partial strong cyclic policy from $s_{10}$ for $A$ and $B$, and can be returned with the flag $\solved$.

\subsubsection{\IDFSRectitle End (Line 34)}

In case none of the actions $a \in \applicableactions(s)$ are able to generate a partial strong cyclic from~$s$ to $A$ and $B$, \IDFSRec returns $\unsolved$.



\SetKwFunction{IDFSSatisf}{\sc idfsp}
\SetKwFunction{IDFSPSearch}{\sc idfsp}
\SetKwFunction{IDFSPSearchRec}{\sc idfsp$_R$}
\SetKwProg{Fn}{}{:}{}

\begin{algorithm}[t!]
\SetInd{0.27em}{0.54em}
\fontsize{8.5}{9.5}\selectfont
\caption{\IDFS Pruning (\IDFSP)}\label{alg:IDFSSatisficing}
\DontPrintSemicolon

\SetKwComment{RefComment}{$\leftarrow$\,}{}

\Fn{\IDFSPSearch{$s_0$}}
{
    $\bound \coloneqq h(s_0), \newbound \coloneqq \infty, \nonpromisingset \coloneqq \emptyset$\;
    \While{$\bound \leq |\S|$}
    {
        $\flag, \pi \coloneqq \IDFSPSearchRec(s_0, \emptyset, \emptyset, \emptypolicy)$\;
        \If{$\flag = \solved$}
        {
            \Return{$\pi$}
        }
        $\bound \coloneqq \newbound, \newbound \coloneqq \infty, \nonpromisingset \coloneqq \emptyset$
    }
    \Return{$\unsolved$}
}

\Fn{\IDFSPSearchRec{$s, \ancestorsset, \ancestorsset_*, \pi$}}
{
    \RefComment{\color{darkgray} Lines 10-13 of Algorithm 1.}

    \If{$s \in \nonpromisingset$}
    {
        \Return{$\unsolved, \pi$}
    }

    $\promising \coloneqq \false$\;
    \For{$a \in \applicableactions(s)$}
    {
        \RefComment{\color{darkgray} Lines 15-20 of Algorithm 1.}
        \tcp{\color{darkgray} Fixed Point.}
            $\ancestorsset_*' \coloneqq \ancestorsset_*, \pi' \coloneqq \pi, \successorsset \coloneqq \succs(s,a), \solvedsuccessorsset \coloneqq \emptyset$\;
        \Repeat{$\reachedfixedpoint$}
        {
            $\reachedfixedpoint \coloneqq \true$\;
            \For{$s' \in (\successorsset \setminus \solvedsuccessorsset) $}
            {
                $\flag, \pi' \coloneqq \IDFSPSearchRec(s', \ancestorsset \cup \lbrace s \rbrace, \ancestorsset_*', \pi')$\;
                \If{$\successorsset  \cap \nonpromisingset \neq \emptyset$}
                {
                    \textbf{break}\;
                }
                \If{$\flag = \solved$}
                    {
                        $\solvedsuccessorsset \coloneqq \solvedsuccessorsset \cup \lbrace s' \rbrace$\;
                        $\ancestorsset_*' \coloneqq \ancestorsset \cup \lbrace s \rbrace$\;
                        $\reachedfixedpoint \coloneqq \false$\;
                    }
            }
            \If{$\successorsset \cap \nonpromisingset \neq \emptyset$}
            {
                \textbf{break}\;
            }
            \If{$\reachedfixedpoint$}
            {
                $\promising \coloneqq \true$\;
            }

        }
        \RefComment{\color{darkgray} Lines 31-33 of Algorithm 1.}
    }
    \If{$\promising = \false$}
    {
        $\nonpromisingset \coloneqq \nonpromisingset \cup \lbrace s \rbrace$\;
    }
    \Return{$\unsolved, \pi$}
}
\end{algorithm}


\subsubsection{\IDFStitle Pruning}

We now present an extended version of \IDFS called \IDFS \textit{Pruning} (\IDFSP). Algorithm~\ref{alg:IDFSSatisficing} presents the pseudo-code of \IDFSP.
In essence, \IDFSP is similar to \IDFS, and the main difference is that it \emph{prunes} states during the searching process. \IDFSPRec considers that a state~$s$ is \emph{promising} if $s \in A$ or at least one of its applicable actions~$a$ reaches the fixed point when evaluating the set of successor states~$\successorsset \coloneqq \succs(s,a)$. If the state~$s$ is not promising, \IDFSPRec adds the state~$s$ into the global set $\nonpromisingset$. \IDFSP sets $\nonpromisingset$ to empty before each iteration.
\IDFSPRec has one additional base case that returns $\unsolved$ if state~$s\in\nonpromisingset$ (Lines~10--11). During the fixed-point computation, \IDFSPRec verifies if at least one of the states in $\successorsset \coloneqq \succs(s,a)$ is in $\nonpromisingset$ and stops the fixed-point computation if it is. This pruning method helps the search because it avoids repeated evaluation of states that generate successors that can not be part of the policy with the current bound.

\section{Minimal Critical-Value in \FONDtitle Planning}\label{sec:critical-path-fond}

\begin{figure}[t]
    \centering
    \begin{subfigure}{.67\columnwidth}
        \centering
        \includegraphics[width=\columnwidth]{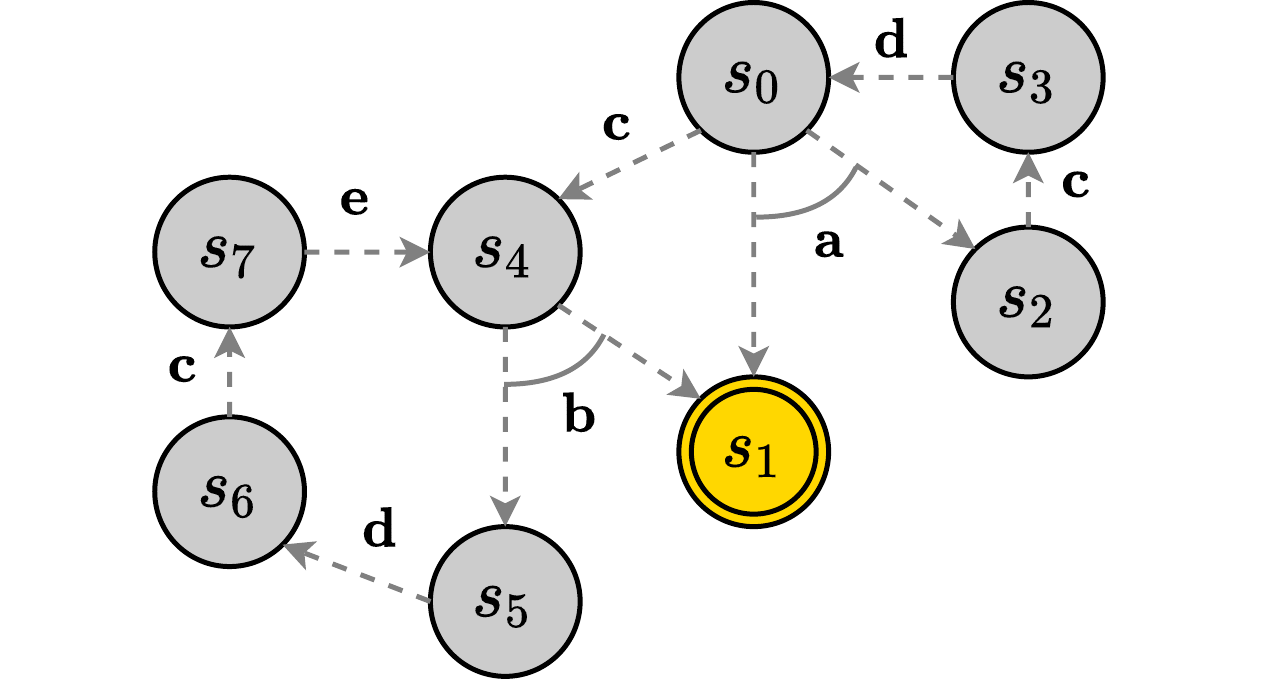}
    \end{subfigure}\\
    \footnotesize (a)\vspace{3mm}\\
    \begin{tabular}[b]{rl}
        \begin{subfigure}{.39\columnwidth}
            \centering
            \includegraphics[width=\columnwidth]{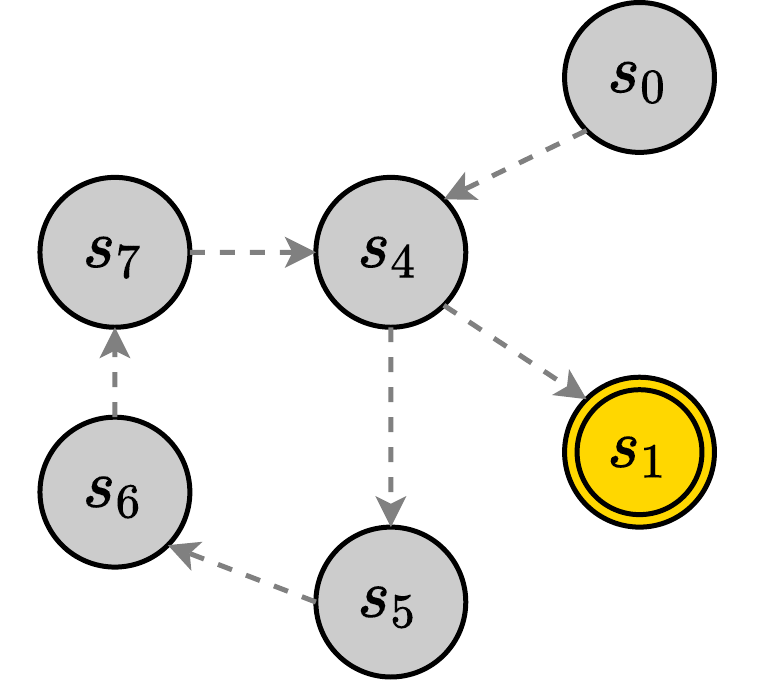}
        \end{subfigure} &
        \begin{subfigure}{.265\columnwidth}
            \centering
            \includegraphics[width=\columnwidth]{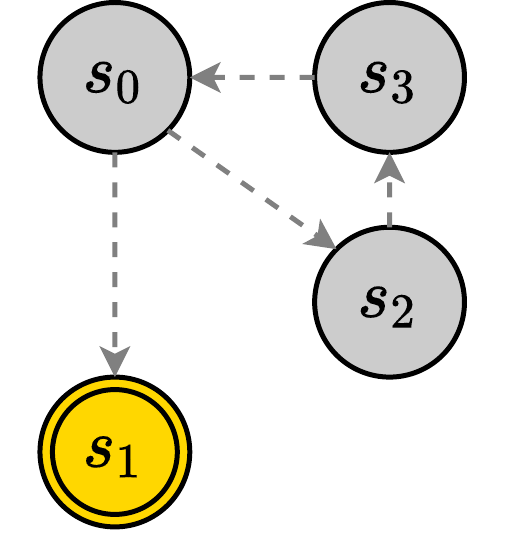}
        \end{subfigure}\vspace{1mm}\\
        \multicolumn{1}{c}{\footnotesize (b)} &
        \multicolumn{1}{c}{\footnotesize (c)} \\
    \end{tabular}
	\caption{\textit{Minimal Critical-Value} example.}
	\label{fig:critical_value_policy}
\end{figure}

We now introduce key properties about the set of strong cyclic policies of a \FOND task~$\Pi$ that are important to characterize the behavior of \IDFS.
A \FOND planning task~$\Pi$ has a set of strong cyclic policies $\P(\Pi)$ -- if $\Pi$ is unsolvable, then $\P(\Pi) = \emptyset$.
Figure~\ref{fig:critical_value_policy}a shows the state-space of a \FOND planning task $\Pi$, with eight states, three deterministic actions, and two non-deterministic actions -- namely~$\{a,b\}$.
This task has only two strong cyclic policies, $\pi_0 = \lbrace s_0 \mapsto \mathbf{c},\break s_4 \mapsto \mathbf{b}, s_5 \mapsto \mathbf{d}, s_6 \mapsto \mathbf{c}, s_7 \mapsto \mathbf{e} \rbrace$ and $\pi_1 = \lbrace s_0 \mapsto \mathbf{a}, s_2 \mapsto \mathbf{c},\break s_3 \mapsto \mathbf{d} \rbrace$.
Figures~\ref{fig:critical_value_policy}b and~\ref{fig:critical_value_policy}c, show respectively the part of the state-space reachable from $s_0$ using each policy.
We use these state-spaces to present the concept of \textit{critical-values} of policies (Definition~\ref{def:st}).

\begin{definition}\label{def:st}
The \textbf{critical-value} $\cv(\pi)$ of a policy $\pi$ is the value of the length of the longest $\pi$-trajectory $\langle s^1, s^2, \dots, s^k \rangle$ with $s^1 = s_0$ and no $i < j \leq k - 1$ with $s^i = s^j$.
\end{definition}

The $\cv$ of $\pi_0$ is generated by $\langle s_0, s_4, s_5, s_6, s_7, s_4 \rangle$, therefore $\cv(\pi_0) = 5$. The $\cv$ of $\pi_1$ is generated by $\langle s_0, s_2, s_3, s_0 \rangle$, therefore $\cv(\pi_1) = 3$.
Definition~\ref{def:cv*} introduces the concept of minimal critical-value $\cv^*$ of a \FOND planning task $\Pi$.

\begin{definition}\label{def:cv*}
The \textbf{minimal critical-value} $\cv^*$ of a \FOND planning task $\Pi$ is equal to $\min_{\pi \in \P(\Pi)} \cv(\pi)$.
\end{definition}

Thus, the $\cv^*$ of the \FOND planning task $\Pi$ in the Figure~\ref{fig:critical_value_policy}a is $\cv^*(\Pi) = \min \lbrace \cv(\pi_0), \cv(\pi_1) \rbrace = \min \lbrace 5, 3 \rbrace = 3$. Definition~\ref{def:cv*} considers all strong cyclic policies of the task~$\Pi$, which are, in general, unavailable. Therefore, we usually do not know the value of $\cv^*(\Pi)$.
Nevertheless, we prove that if \IDFS uses $\minF$ and an admissible heuristic function for the deterministic version of the task~$\Pi$, \IDFS will search to a depth of at most $\cv^*$. Thus, if~$\Pi$ is \textit{solvable}, \IDFS will return a strong cyclic policy before the search starts evaluating states at a depth greater than $\cv^*$.

\aboverulesep = 0.15mm
\belowrulesep = 0.3mm

\begin{table*}[h!]
\centering
\fontsize{9}{10}\selectfont

\begin{tabular}{rrrrrrXrrrrr}
\toprule
 & \multicolumn{5}{c}{\IDFS~($\minF$, $\BLIND$)} & & \multicolumn{5}{c}{\IDFS~($\minF$, $\hMAX$)} \\

\cmidrule[\heavyrulewidth]{2-6} \cmidrule[\heavyrulewidth]{8-12}

\textbf{Domain (\#)}
& \multicolumn{1}{r}{\it C} & \multicolumn{1}{r}{\it T} & \multicolumn{1}{r}{$|\pi|$} & \multicolumn{1}{r}{$b_I$/$b_F$} & \multicolumn{1}{r}{$i$} & \multicolumn{1}{c}{}
& \multicolumn{1}{r}{\it C} & \multicolumn{1}{r}{\it T} & \multicolumn{1}{r}{$|\pi|$} & \multicolumn{1}{r}{$b_I$/$b_F$} & \multicolumn{1}{r}{$i$} \\ \midrule

\rowcolor{grayline}\doors (\#15) & 11 & 30.1 & 1486.7 & 0.0/8.0 & 8.0 & & 11 & 10.9 & 1486.7 & 7.0/8.0 & 2.0 \\
\islands (\#60) & 29 & 18.7 & 4.9 & 0.0/4.9 & 4.9 & & 60 & 0.1 & 4.9 & 4.9/4.9 & 1.0 \\
\rowcolor{grayline}\miner (\#51) & 0 & - & - & -/- & - & & 40 & - & - & -/- & - \\
\twspiky (\#11) & 4 & 18.5 & 26.0 & 0.0/22.0 & 22.0 & & 9 & 3.7 & 25.0 & 8.0/22.0 & 15.0 \\
\rowcolor{grayline}\twtruck (\#74) & 13 & 23.4 & 13.8 & 0.0/10.8 & 10.8 & & 26 & 0.6 & 13.2 & 4.2/10.8 & 7.7 \\

\midrule
\textbf{Sub-Total (\#211)} & 57 & 22.6 & 382.5 & 0.0/11.4 & 11.4 & & 146 & 3.8 & 382.4 & 6.1/11.4 & 6.4 \\
\midrule

\acrobatics (\#8) & 4 & 2.3 & 14.0 & 0.0/14.0 & 14.0 & & 8 & 0.1 & 14.0 & 3.8/14.0 & 11.3 \\
\rowcolor{grayline}\beamwalk (\#11) & 8 & 29.2 & 254.0 & 0.0/254.0 & 254.0 & & 8 & 9.5 & 254.0 & 127.5/254.0 & 127.5 \\
\bworiginal (\#30) & 10 & 17.4 & 13.5 & 0.0/7.5 & 7.5 & & 10 & 4.2 & 12.4 & 2.8/7.5 & 5.7 \\
\rowcolor{grayline}\bwtwo (\#15) & 5 & 38.0 & 14.4 & 0.0/9.4 & 9.4 & & 5 & 4.9 & 14.2 & 2.8/9.4 & 7.6 \\
\bwnew (\#40) & 6 & 26.8 & 8.0 & 0.0/5.5 & 5.5 & & 6 & 2.5 & 8.0 & 2.2/5.5 & 4.2 \\
\rowcolor{grayline}\chain (\#10) & 2 & 62.8 & 42.0 & 0.0/28.0 & 28.0 & & 10 & 0.1 & 42.0 & 28.0/28.0 & 1.0 \\
\earthobs (\#40) & 8 & 3.9 & 19.3 & 0.0/9.6 & 9.6 & & 9 & 0.4 & 18.3 & 4.4/9.6 & 6.0 \\
\rowcolor{grayline}\elevators (\#15) & 4 & 44.0 & 12.0 & 0.0/11.3 & 11.3 & & 5 & 1.5 & 11.3 & 4.8/11.3 & 7.5 \\
\faults (\#55) & 18 & 14.8 & 28.4 & 0.0/7.3 & 7.3 & & 19 & 7.7 & 21.6 & 2.0/7.3 & 6.3 \\
\rowcolor{grayline}\firstresp (\#100) & 20 & 22.5 & 5.7 & 0.0/5.7 & 5.7 & & 23 & 3.7 & 6.3 & 2.6/5.7 & 4.0 \\
\tritw (\#40) & 3 & 23.2 & 22.0 & 0.0/15.0 & 15.0 & & 3 & 13.4 & 22.0 & 4.0/15.0 & 12.0 \\
\rowcolor{grayline}\zeno (\#15) & 0 & - & - & -/- & - & & 3 & - & - & -/- & - \\
\midrule
\textbf{Total (\#590)} & 145 & 25.0 & 131.0 & 0.0/27.5 & 27.5 & & 255 & 3.7 & 130.3 & 13.9/27.5 & 14.6 \\

\bottomrule
\end{tabular}

\caption{\IDFS comparison with $\minF$: $\BLIND$ vs $\hMAX$.}
\label{tab:results-admissible}
\end{table*}


\section{Theoretical Properties}\label{sec:theory}

In this section, we present a proof idea that shows that if a \FOND planning task~$\Pi$ is \emph{solvable}, \IDFS returns a \emph{strong cyclic policy} by searching to a depth of at most~$\cv^*(\Pi)$, and if $\Pi$ is \emph{unsolvable}, \IDFS identifies it correctly.
Theorem~\ref{theorem} bounds the behavior of the \IDFS algorithm  
by the structure of the \FOND planning task~$\Pi$.

\begin{theorem}~\label{theorem}
Given a \FOND planning task~$\Pi$, an admissible heuristic function~$h$ for a \emph{deterministic} version of~$\Pi$, and \IDFS using $\F_{\min}$. If~$\Pi$ is \emph{solvable}, then \IDFS returns a \emph{strong cyclic policy}~$\pi$ by searching to a depth of at most~$\cv^*(\Pi)$. If $\Pi$ is \emph{unsolvable}, then \IDFS returns $\unsolvable$.
\end{theorem}
\emph{Proof Idea.} If~$\Pi$ is \emph{solvable}, then there is a strong cyclic policy~$\pi$ which has $\cv(\pi)=\cv^*(\Pi)$.
Suppose state~$s$ is part of the policy~$\pi$, \IDFSSearchRec analyzes all actions applicable on~$s$, including the action that is part of the policy~$\pi(s)$, with incremental search depths and using $\F_{\min}$ and heuristic~$h$ when possible.
Since~$s$ is in the policy~$\pi$, \IDFSSearchRec can, by the construction of the algorithm, find a \emph{policy} that includes~$s$ searching to a depth of at most~$\cv^{*}$.
Because task~$\Pi$ is \emph{solvable} and $s_0$ is in any policy including policy~$\pi$, \IDFS returns a \emph{strong cyclic policy} by searching to a depth of at most~$\cv^{*}$.
\IDFSSearchRec only returns $\solved$ for a state $s$ using action~$a$ if all its successors in~$\succs(s,a)$ return $\solved$. Thus, if a \FOND planning task~$\Pi$ is \emph{unsolvable}, \IDFS returns $\unsolvable$.
\IDFS always terminates because the state-space size limits the number of iterations of the main loop.

\section{Experiments and Evaluation}\label{sec:experiments}

We now present the set of experiments we have conducted to evaluate the efficiency of our \IDFS algorithm for solving \FOND planning tasks.
We compare our algorithm to state-of-the-art \FOND planners, such as \PRP~\cite{Muise12ICAPSFond}, \MYND~\cite{MyND_MattmullerOHB10}, and \FONDSAT~\cite{GeffnerG18_FONDSAT}. We have implemented our algorithm using part of the source code of \MYND.
We use the delete relaxation heuristic functions for the deterministic version of the planning task as proposed by~\citet{Mattmueller_thesis_2013}.
As a result, we have a \FOND planner called \Paladinus\footnote{\Paladinus code: \url{https://github.com/ramonpereira/paladinus}}.

We empirically evaluate \IDFS using two distinct benchmark sets: IPC-\FOND and NEW-\FOND. IPC-\FOND contains 379 planning tasks over 12 \FOND domains from the IPC \shortcite{IPC6}~and~\cite{Muise12ICAPSFond}.
The NEW-\FOND benchmark set~\cite{GeffnerG18_FONDSAT} introduces \FOND planning tasks that contain several trajectories to goal states that are not part of any strong cyclic policy.
NEW-\FOND contains 211 tasks over five \FOND domains, namely \doors, \islands, \miner, \twspiky, and \twtruck. Note that 25 out of 590 tasks are \emph{unsolvable}, namely, 25 \FOND planning tasks of \firstresp -- a domain of IPC-\FOND.

We have run all experiments using a single core of a 12 core Intel(R) Xeon(R) CPU E5-2620 v3 @ 2.40GHz with 16GB of RAM, with a memory limit of 4GB, and set a 5 minute (300 seconds) time-out per planning task.
We evaluate the planners, when applicable, using the following metrics: \emph{number of solved tasks}, i.e., \emph{coverage} ($C$), \emph{time to solve} ($T$) in seconds, \emph{average policy size} ($|\pi|$), \emph{initial bound} and the \emph{final bound} (respectively, $b_I$ and $b_F$), and the \emph{number iterations} ($i$).
Apart from the coverage ($C$), all results shown in Tables~\ref{tab:results-admissible},~\ref{tab:results-variations-idfs-1},~\ref{tab:results-variations-idfs-2}, and~\ref{tab:results-comparsion} are calculated over the intersection of the tasks solved by all planners in the respective table.

\subsubsection{\IDFStitle with Admissible Heuristic Functions}

\begin{table*}[t!]
    \centering
    \fontsize{9}{10}\selectfont

    \begin{tabular}{rrrrrrXrrrrr}
    \toprule
     & \multicolumn{5}{c}{\IDFS~($\minF$, $\hADD$)}
     & & \multicolumn{5}{c}{\IDFS~($\maxF$, $\hADD$)} \\

    \cmidrule[\heavyrulewidth]{2-6} \cmidrule[\heavyrulewidth]{8-12}

    \textbf{Domain (\#)}
    & \multicolumn{1}{r}{\it C} & \multicolumn{1}{r}{\it T} & \multicolumn{1}{r}{$|\pi|$} & \multicolumn{1}{r}{$b_I$/$b_F$} & \multicolumn{1}{r}{$i$} &
    & \multicolumn{1}{r}{\it C} & \multicolumn{1}{r}{\it T} & \multicolumn{1}{r}{$|\pi|$} & \multicolumn{1}{r}{$b_I$/$b_F$} & \multicolumn{1}{r}{$i$} \\ \midrule

    \rowcolor{grayline}\doors (\#15) & 11 & 14.1 & 1486.7 & 26.7/35.7 & 1.9 & & 11 & 10.7 & 1486.7 & 26.7/120.5 & 2.8 \\
    \islands (\#60) & 60 & 0.5 & 7.0 & 7.0/7.0 & 1.0 & & 60 & 0.6 & 7.0 & 7.0/7.0 & 1.0 \\
    \rowcolor{grayline}\miner (\#51) & 51 & 1.1 & 23.2 & 39.6/39.9 & 1.2 & & 51 & 1.8 & 23.2 & 39.6/39.9 & 1.2 \\
    \twspiky (\#11) & 9 & 14.9 & 25.0 & 8.0/22.0 & 15.0 & & 6 & 39.8 & 25.0 & 8.0/24.0 & 17.0 \\
    \rowcolor{grayline}\twtruck (\#74) & 26 & 19.3 & 14.3 & 3.9/11.1 & 8.2 & & 21 & 30.0 & 14.4 & 3.9/12.1 & 9.2 \\
    
    \midrule
    \textbf{Sub-Total (\#211)} & 157 & 9.9 & 311.2 & 17.1/23.1 & 5.4 & & 149 & 16.6 & 311.2 & 17.1/40.7 & 6.2 \\
    \midrule

    \acrobatics (\#8) & 8 & 1.6 & 126.5 & 63.8/126.5 & 63.8 & & 8 & 0.6 & 126.5 & 63.8/748.1 & 73.9 \\
    \rowcolor{grayline}\beamwalk (\#11) & 11 & 0.5 & 453.2 & 453.2/453.2 & 1.0 & & 9 & 12.0 & 453.2 & 453.2/39176.7 & 113.6 \\
    \bworiginal (\#30) & 15 & 10.5 & 23.2 & 14.6/15.2 & 1.6 & & 25 & 0.6 & 17.3 & 14.6/22.4 & 2.0 \\
    \rowcolor{grayline}\bwtwo (\#15) & 7 & 1.6 & 21.6 & 15.4/17.0 & 2.1 & & 14 & 0.8 & 21.3 & 15.4/22.6 & 2.0 \\
    \bwnew (\#40) & 10 & 3.3 & 20.0 & 12.3/12.8 & 1.4 & & 19 & 0.6 & 17.4 & 12.3/19.3 & 2.0 \\
    \rowcolor{grayline}\chain (\#10) & 10 & 0.3 & 162.0 & 161.0/161.8 & 1.8 & & 10 & 0.3 & 162.0 & 161.0/161.8 & 1.8 \\
    \earthobs (\#40) & 18 & 0.2 & 47.4 & 22.9/23.9 & 1.7 & & 19 & 0.5 & 38.8 & 22.9/25.8 & 2.5 \\
    \rowcolor{grayline}\elevators (\#15) & 10 & 0.1 & 19.6 & 21.1/21.7 & 1.6 & & 8 & 0.2 & 19.6 & 21.1/21.9 & 1.7 \\
    \faults (\#55) & 22 & 19.5 & 26.8 & 5.9/7.9 & 3.0 & & 23 & 16.2 & 26.8 & 5.9/9.1 & 2.6 \\
    \rowcolor{grayline}\firstresp (\#100) & 57 & 0.3 & 13.7 & 12.6/12.6 & 1.0 & & 31 & 28.7 & 14.3 & 12.6/13.8 & 2.2 \\
    \tritw (\#40) & 3 & 13.2 & 22.0 & 4.0/15.0 & 12.0 & & 3 & 9.8 & 22.0 & 4.0/15.0 & 8.0 \\
    \rowcolor{grayline}\zeno (\#15) & 7 & 13.4 & 29.5 & 30.0/31.2 & 2.2 & & 6 & 17.8 & 29.5 & 30.0/31.2 & 2.2 \\
    \midrule

    \textbf{Total (\#590)} & 335 & 6.7 & 148.3 & 53.1/59.7 & 7.1 & & 324 & 10.1 & 147.4 & 53.1/2380.7 & 14.5 \\

    \bottomrule
    \end{tabular}

    \caption{\IDFS algorithm using $\hADD$ with $\minF$ and $\maxF$, without pruning.}
    \label{tab:results-variations-idfs-1}
    \end{table*}

    \begin{table*}[t!]
    \centering
    \fontsize{9}{10}\selectfont

    \begin{tabular}{rrrrrrXrrrrr}
    \toprule
     & \multicolumn{5}{c}{\IDFSP~($\minF$, $\hADD$)}
     & & \multicolumn{5}{c}{\IDFSP~($\maxF$, $\hADD$)}  \\

    \cmidrule[\heavyrulewidth]{2-6} \cmidrule[\heavyrulewidth]{8-12}

    \textbf{Domain (\#)}
    & \multicolumn{1}{r}{\it C} & \multicolumn{1}{r}{\it T} & \multicolumn{1}{r}{$|\pi|$} & \multicolumn{1}{r}{$b_I$/$b_F$} & \multicolumn{1}{r}{$i$} &
    & \multicolumn{1}{r}{\it C} & \multicolumn{1}{r}{\it T} & \multicolumn{1}{r}{$|\pi|$} & \multicolumn{1}{r}{$b_I$/$b_F$} & \multicolumn{1}{r}{$i$} \\ \midrule

    \rowcolor{grayline}\doors (\#15) & 13 & 2.3 & 1486.7 & 26.7/35.7 & 1.9 & & 13 & 1.7 & 1486.7 & 26.7/120.5 & 2.8 \\
    \islands (\#60) & 60 & 0.3 & 7.0 & 7.0/7.0 & 1.0 & & 60 & 0.5 & 7.0 & 7.0/7.0 & 1.0 \\
    \rowcolor{grayline}\miner (\#51) & 51 & 0.8 & 23.2 & 39.6/39.9 & 1.2 & & 51 & 0.9 & 23.2 & 39.6/39.9 & 1.2 \\
    \twspiky (\#11) & 10 & 2.5 & 1409.3 & 8.0/20.0 & 13.0 & & 10 & 3.4 & 1409.3 & 8.0/22.0 & 15.0 \\
    \rowcolor{grayline}\twtruck (\#74) & 55 & 0.2 & 17.4 & 3.9/12.8 & 9.9 & & 44 & 2.1 & 19.8 & 3.9/17.3 & 14.4 \\ 
    
    \midrule
    \textbf{Sub-Total (\#211)} & 189 & 1.2 & 588.7 & 17.1/23.1 & 5.4 & & 178 & 1.7 & 589.2 & 17.1/41.3 & 6.8 \\
    \midrule

    \acrobatics (\#8) & 8 & 0.1 & 126.5 & 63.8/63.8 & 1.0 & & 8 & 0.6 & 126.5 & 63.8/749.8 & 75.5 \\
    \rowcolor{grayline}\beamwalk (\#11) & 11 & 0.4 & 453.2 & 453.2/453.2 & 1.0 & & 11 & 0.5 & 453.2 & 453.2/39176.7 & 113.6 \\
    \bworiginal (\#30) & 16 & 7.5 & 23.8 & 14.6/17.6 & 3.9 & & 29 & 0.3 & 16.9 & 14.6/22.6 & 2.2 \\
    \rowcolor{grayline}\bwtwo (\#15) & 10 & 1.0 & 17.4 & 15.4/18.6 & 3.4 & & 15 & 0.3 & 19.1 & 15.4/22.9 & 2.3 \\
    \bwnew (\#40) & 12 & 1.6 & 14.4 & 12.3/14.9 & 3.5 & & 21 & 0.2 & 17.4 & 12.3/19.3 & 2.0 \\
    \rowcolor{grayline}\chain (\#10) & 10 & 0.3 & 162.0 & 161.0/161.8 & 1.8 & & 10 & 0.3 & 162.0 & 161.0/161.8 & 1.8 \\
    \earthobs (\#40) & 19 & 1.4 & 40.9 & 22.9/28.1 & 4.3 & & 25 & 0.1 & 35.6 & 22.9/27.9 & 3.9 \\
    \rowcolor{grayline}\elevators (\#15) & 9 & 0.1 & 19.4 & 21.1/21.7 & 1.6 & & 8 & 0.1 & 19.4 & 21.1/21.9 & 1.7 \\
    \faults (\#55) & 55 & 0.1 & 47.5 & 5.9/7.5 & 1.8 & & 55 & 0.1 & 43.1 & 5.9/8.9 & 2.3 \\
    \rowcolor{grayline}\firstresp (\#100) & 60 & 0.3 & 19.7 & 12.6/12.6 & 1.1 & & 46 & 5.9 & 109.5 & 12.6/13.8 & 2.3 \\
    \tritw (\#40) & 36 & 0.1 & 26.0 & 4.0/13.3 & 10.3 & & 8 & 0.1 & 22.0 & 4.0/15.0 & 8.0 \\
    \rowcolor{grayline}\zeno (\#15) & 6 & 12.8 & 29.7 & 30.0/31.2 & 2.2 & & 8 & 12.9 & 29.7 & 30.0/31.2 & 2.2 \\
    \midrule

    \textbf{Total (\#590)} & 411 & 1.9 & 230.8 & 53.1/56.4 & 3.7 & & 422 & 1.8 & 235.3 & 53.1/2381.1 & 14.8 \\

    \bottomrule
    \end{tabular}

    \caption{\IDFS algorithm using $\hADD$ with $\minF$ and $\maxF$, with pruning.}
    \label{tab:results-variations-idfs-2}
    \end{table*}

\newlength\empiricalfiguressize
\setlength\empiricalfiguressize{0.8\columnwidth}

\begin{figure*}[h!]
	\centering
	\begin{subfigure}[b]{\empiricalfiguressize}
 	    \includegraphics[width=\columnwidth]{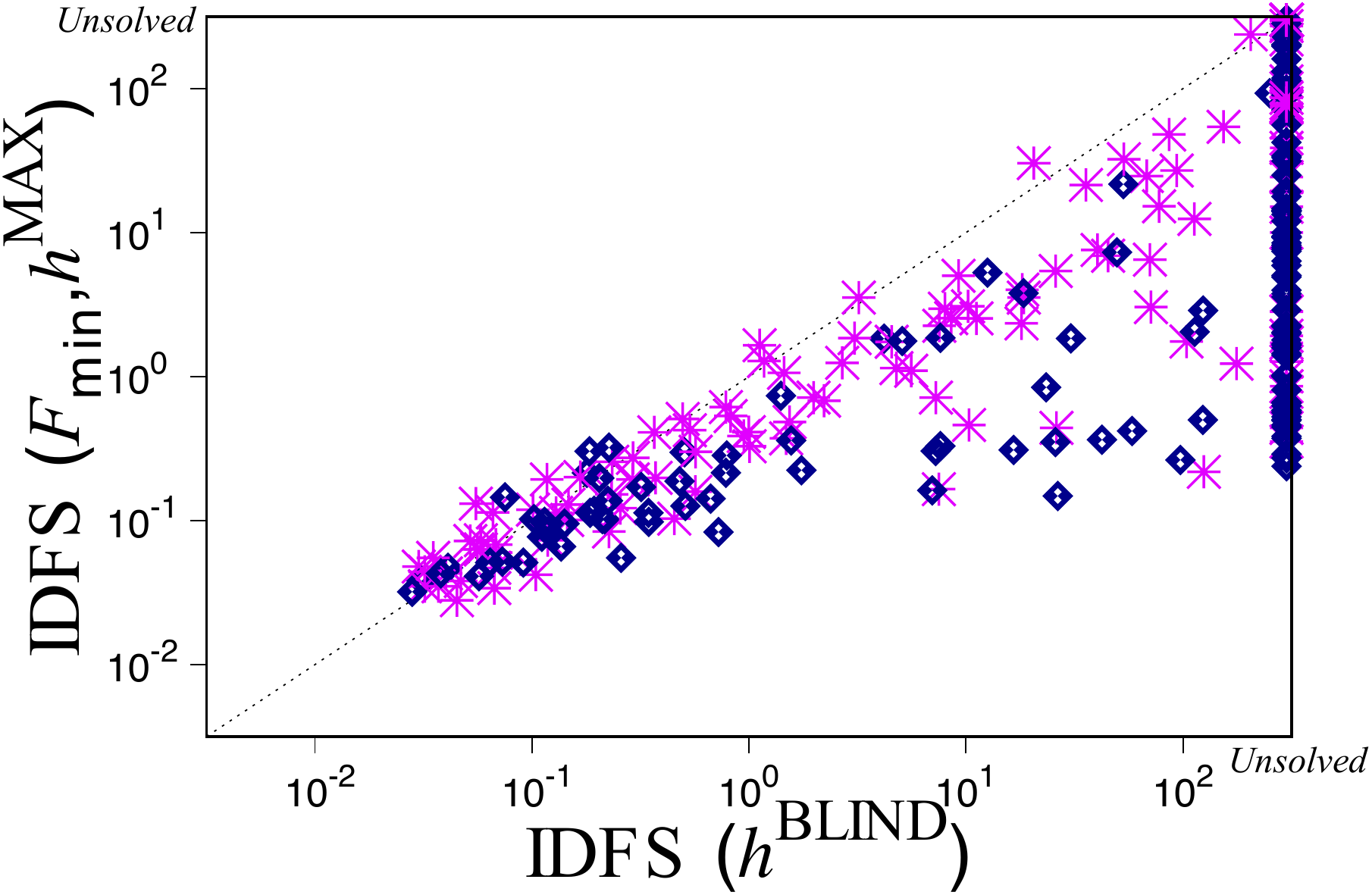}
		\caption{}
		\label{fig:max_blind}
	\end{subfigure}
	\hspace{8mm}
	\begin{subfigure}[b]{\empiricalfiguressize}
		\includegraphics[width=\columnwidth]{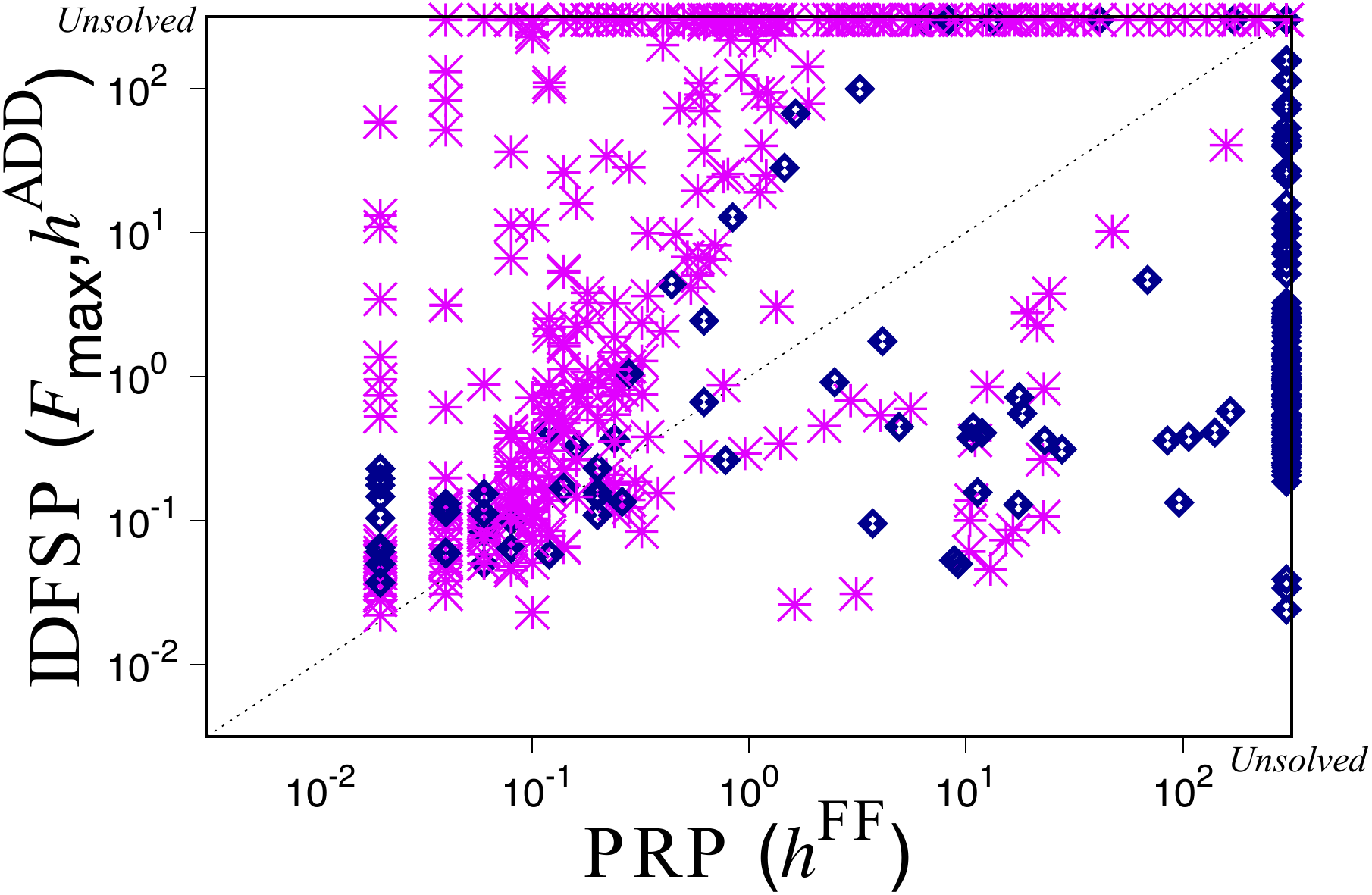}
		\caption{}
		\label{fig:idfsbest_prp}
	\end{subfigure} \\
	\begin{subfigure}[b]{\empiricalfiguressize}
		\includegraphics[width=\columnwidth]{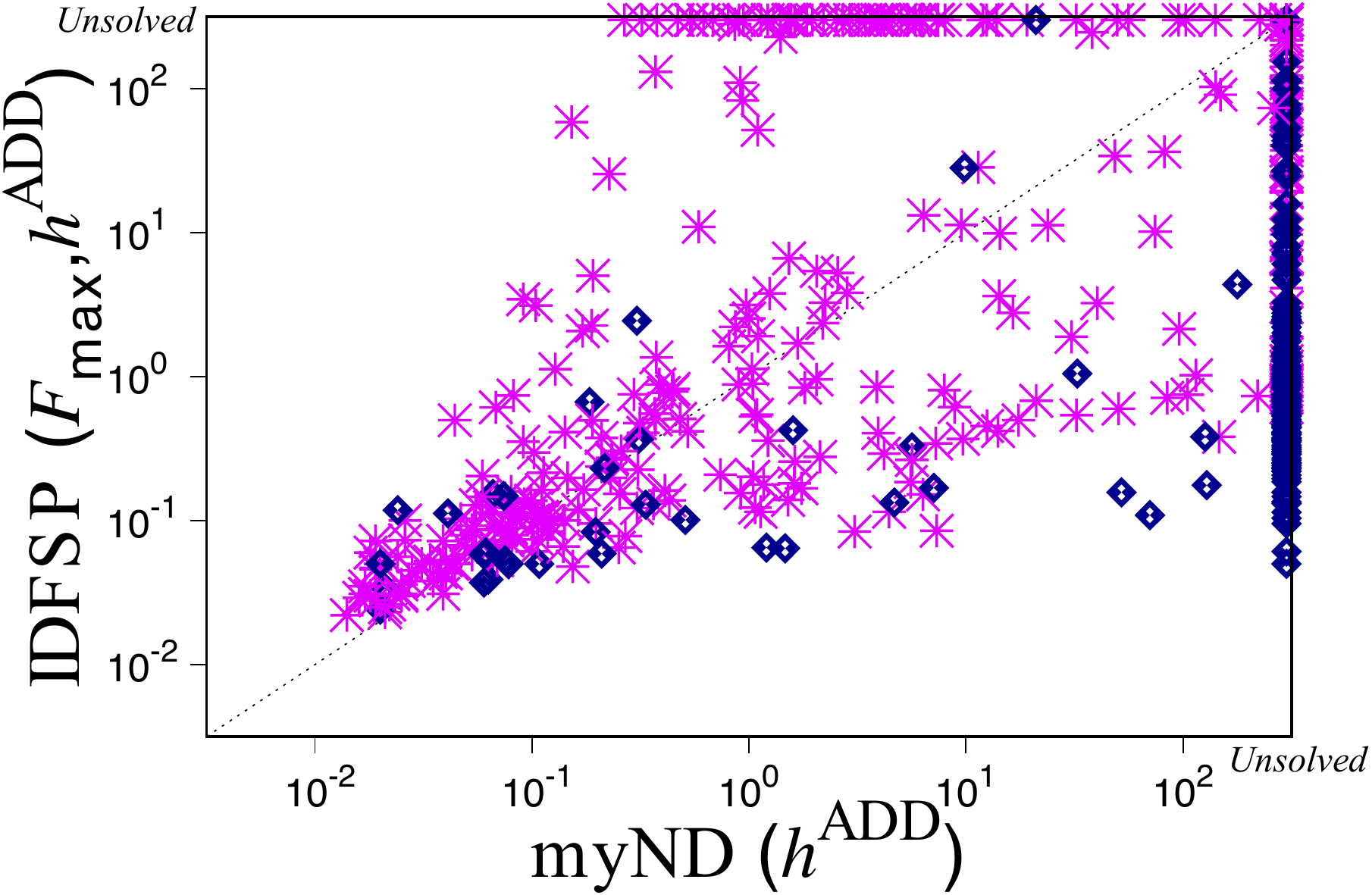}
		\caption{}
		\label{fig:idfsbest_mynd}
	\end{subfigure}
	\hspace{8mm}
	\begin{subfigure}[b]{\empiricalfiguressize}
		\includegraphics[width=\columnwidth]{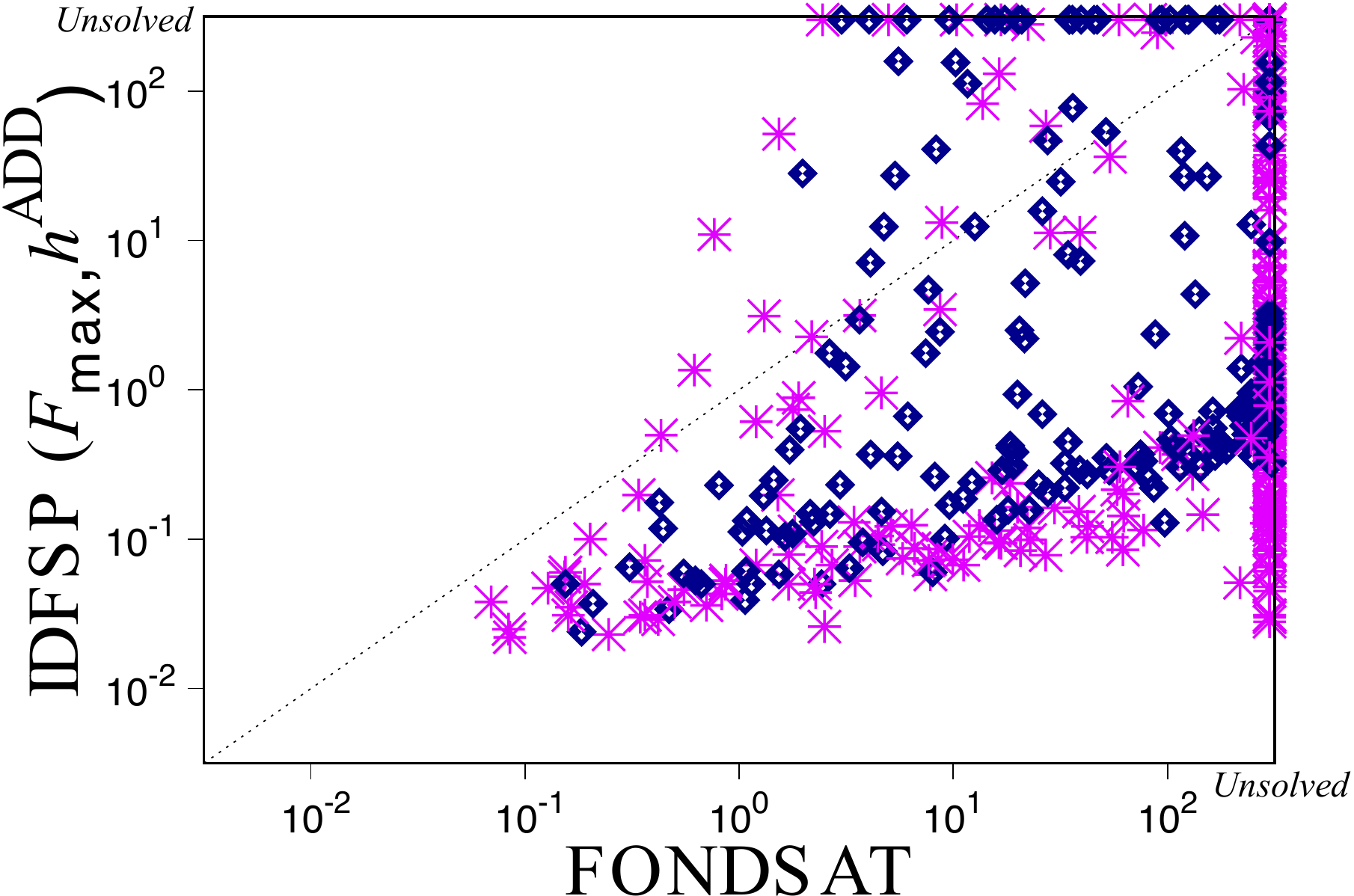}
		\caption{}
		\label{fig:idfsbest_fondsat}
	\end{subfigure}

	\caption{Planning time (in seconds) per planning task. Blue diamonds: results for the NEW-\FOND benchmark set; Dark-pink asterisks: results for the IPC-\FOND benchmark set.}
	\label{fig:planning-time}
\end{figure*}



We start our evaluation by presenting a comparison of \IDFS using $\BLIND$ and $\hMAX$ with the evaluation function $\minF$. This comparison evaluates how useful the information of the heuristic function is for \IDFS concerning search efficiency.
We evaluate this with the following metrics: the number of solved tasks, the time to solve, and the number of iterations required to solve the task -- fewer iterations mean that \IDFS reaches faster the depth where it finds a strong cyclic policy.
Table~\ref{tab:results-admissible} summarizes the results for all 17 \FOND domains of the used benchmark sets, showing the performance of \IDFS when using $\minF$ with $\hMAX$ and $\BLIND$, denoted as \IDFS($\minF$,~$\hMAX$) and \IDFS($\minF$,~$\BLIND$), respectively.

\IDFS($\minF$,~$\hMAX$) solves in total 255 tasks, whereas \IDFS($\minF$,~$\BLIND$) solves 145 tasks. Both \IDFS($\minF$, $\hMAX$) and \IDFS($\minF$,~$\BLIND$) identified the 25 tasks of \firstresp as \emph{unsolvable}. \IDFS($\minF$,~$\BLIND$) exceeded the time limit to solve all tasks of \miner and \zeno. Table~\ref{tab:results-admissible} shows that \IDFS($\minF$, $\hMAX$) always uses fewer iterations to solve the same tasks when compared to \IDFS($\minF$,~$\BLIND$), and it also shows that, in general, \IDFS($\minF$, $\hMAX$) is much faster even considering the cost of computing the heuristic function.

Figure~3a shows the planning time comparison between \IDFS($\minF$, $\hMAX$) and \IDFS($\minF$,~$\BLIND$). Overall, \IDFS($\minF$, $\hMAX$) outperforms \IDFS($\minF$,~$\BLIND$) with respect to planning time among most planning tasks, especially over the NEW-\FOND benchmarks (blue diamond in Figure~3a).
Thus, we conclude that, in general, \IDFS benefits from using the information of the heuristic function.

\subsubsection{\IDFStitle vs. \IDFStitle Pruning}

We now evaluate our \IDFS algorithm using $\hADD$ with $\maxF$ and $\minF$. We also compare the versions of \IDFS with and without pruning.
Tables~\ref{tab:results-variations-idfs-1} and~\ref{tab:results-variations-idfs-2} show the results the four variations of \IDFS with $\hADD$.
Note that all four variations of \IDFS with $\hADD$ solved more tasks than both \IDFS($\minF$, $\hMAX$) and \IDFS($\minF$, $\BLIND$). Such empirical results show that using a more informative heuristic has a significant impact on the results. \IDFS($\minF$, $\hADD$) solved 80 tasks more than \IDFS($\minF$, $\hMAX$). \IDFSP($\maxF$, $\hADD$) outperforms the other variants in terms of coverage and planning time. However, the average final bound~$b_F$, and the average number of iterations~$i$ for the intersection of the solved tasks are higher for variations with $\maxF$ compared to the variations with $\minF$.
Also, the pruning variants (\IDFSP) are far superior to the variants without pruning.

\begin{table}[t!]
\centering
\fontsize{9}{9}\selectfont

\begin{tabular}{lr}
\toprule

\textbf{Planner} & \textit{Solved Tasks (\#590)} \\

\midrule

\rowcolor{grayline}\Paladinus~\IDFSP~($\minF$, $\hMAX$) & 337 \\
\Paladinus~\IDFSP~($\minF$, $\hFF$) & 406 \\
\rowcolor{grayline}\Paladinus~\IDFSP~($\minF$, $\hADD$) & 411 \\

\midrule

\Paladinus~\IDFSP~($\maxF$, $\hMAX$) & 334 \\
\rowcolor{grayline}\Paladinus~\IDFSP~($\maxF$, $\hFF$) & 380 \\
\Paladinus~\IDFSP~($\maxF$, $\hADD$) & \textbf{422} \\

\midrule

\rowcolor{grayline}\FONDSAT & 276 \\

\midrule

\PRP~($\hMAX$) & 292 \\
\rowcolor{grayline}\PRP~($\hFF$) & 412 \\
\PRP~($\hADD$) & 389 \\

\midrule

\rowcolor{grayline}\MYND~($\hMAX$) & 180 \\
\MYND~($\hFF$) & 265 \\
\rowcolor{grayline}\MYND~($\hADD$) & 289 \\

\bottomrule
\end{tabular}

\caption{Overall coverage results. }
\label{tab:avg_results-comparsion}
\end{table}

\subsubsection{Comparison with other \FONDtitle Planners}

\begin{table*}[t!]
\centering
\fontsize{9}{10}\selectfont

\begin{tabular}{rrrrXrrrXrrrXrrr}
\toprule
 & \multicolumn{3}{c}{\IDFSSatisficing~($\maxF$, $\hADD$)}
 & & \multicolumn{3}{c}{\PRP ($\hFF$)}
 & & \multicolumn{3}{c}{\MYND ($\hADD$)}
 & & \multicolumn{3}{c}{\FONDSAT} \\

\cmidrule[\heavyrulewidth]{2-4} \cmidrule[\heavyrulewidth]{6-8}
\cmidrule[\heavyrulewidth]{10-12}
\cmidrule[\heavyrulewidth]{14-16}

\textbf{Domain (\#)}
& \multicolumn{1}{r}{\it C} & \multicolumn{1}{r}{\it T} & \multicolumn{1}{r}{$|\pi|$} & \multicolumn{1}{r}{}
& \multicolumn{1}{r}{\it C} & \multicolumn{1}{r}{\it T} & \multicolumn{1}{r}{$|\pi|$} & \multicolumn{1}{r}{}
& \multicolumn{1}{r}{\it C} & \multicolumn{1}{r}{\it T} & \multicolumn{1}{r}{$|\pi|$} & \multicolumn{1}{r}{}
& \multicolumn{1}{r}{\it C} & \multicolumn{1}{r}{\it T} & \multicolumn{1}{r}{$|\pi|$} \\ \midrule

\rowcolor{grayline}\doors (\#15) & 13 & 0.34 & 670.0 & & 12 & 0.13 & 16.0 & & 9 & 6.77 & 670.0 & & 10 & 23.48 & 16.0 \\
\islands (\#60) & 60 & 0.10 & 6.5 & & 27 & 0.08 & 7.5 & & 12 & 11.06 & 6.83 & & 46 & 4.38 & 7.5 \\
\rowcolor{grayline}\miner (\#51) & 51 & - & - & & 9 & - & - & & 0 & - & - & & 28 & - & - \\
\twspiky (\#11) & 10 & 0.13 & 25.0 & & 1 & 17.4 & 23.0 & & 1 & 0.33 & 25.0 & & 3 & 97.07 & 23.0 \\
\rowcolor{grayline}\twtruck (\#74) & 44 & 2.97 & 21.27 & & 17 & 20.34 & 19.36 & & 12 & 12.94 & 13.82 & & 67 & 4.51 & 12.18 \\

\midrule
\textbf{Sub-Total (\#211)} & 178 & 0.88 & 180.69 & & 66 & 9.49 & 16.47 & & 36 & 7.77 & 178.91 & & 154 & 32.36 & 14.67 \\
\midrule

\acrobatics (\#8) & 8 & 0.05 & 8.33 & & 8 & 9.43 & 9.33 & & 8 & 0.02 & 8.33 & & 3 & 3.04 & 9.33 \\
\rowcolor{grayline}\beamwalk (\#11) & 11 & 0.02 & 11.0 & & 11 & 0.86 & 12.0 & & 10 & 0.02 & 11.0 & & 2 & 1.37 & 12.0 \\
\bworiginal (\#30) & 29 & 0.10 & 12.2 & & 30 & 0.06 & 11.7 & & 15 & 0.10 & 11.6 & & 10 & 15.02 & 11.1 \\
\rowcolor{grayline}\bwtwo (\#15) & 15 & 0.12 & 13.2 & & 15 & 0.08 & 14.4 & & 6 & 0.23 & 17.6 & & 5 & 24.71 & 12.2 \\
\bwnew (\#40) & 21 & 0.08 & 8.33 & & 40 & 0.05 & 7.83 & & 9 & 0.08 & 8.5 & & 6 & 14.85 & 7.5 \\
\rowcolor{grayline}\chain (\#10) & 10 & 0.05 & 27.0 & & 10 & 0.1 & 28.0 & & 10 & 0.07 & 27.0 & & 1 & 218.39 & 28.0 \\
\earthobs (\#40) & 25 & - & - & & 40 & - & - & & 25 & - & - & & 0 & - & - \\
\rowcolor{grayline}\elevators (\#15) & 8 & 0.07 & 19.43 & & 15 & 0.05 & 17.71 & & 10 & 1.11 & 18.57 & & 7 & 19.01 & 15.86 \\
\faults (\#55) & 55 & 0.14 & 120.66 & & 55 & 0.06 & 11.48 & & 53 & 0.95 & 67.55 & & 29 & 38.05 & 11.48 \\
\rowcolor{grayline}\firstresp (\#100) & 46 & 34.68 & 103.16 & & 75 & 0.62 & 10.22 & & 58 & 8.65 & 10.95 & & 44 & 27.86 & 9.57 \\
\tritw (\#40) & 8 & 0.08 & 22.0 & & 32 & 0.1 & 23.0 & & 40 & 0.04 & 34.0 & & 3 & 51.42 & 16.0 \\
\rowcolor{grayline}\zeno (\#15) & 8 & 1.01 & 27.0 & & 15 & 0.13 & 23.67 & & 5 & 0.44 & 22.67 & & 3 & 137.64 & 16.33 \\ \midrule

\textbf{Total (\#590)} & 422 & 2.38 & 93.12 & & 412 & 2.91 & 13.84 & & 289 & 4.77 & 90.17 & & 276 & 45.22 & 13.03 \\

\bottomrule
\end{tabular}

\caption{Comparison with \PRP, \MYND, and \FONDSAT.}
\label{tab:results-comparsion}
\end{table*}

Finally, we conclude our evaluation by comparing the best variation of our algorithm (\IDFSP($\maxF$, $\hADD$)) with the state-of-the-art in \FOND planning, i.e., the \PRP, \MYND, and \FONDSAT planners.
Table~\ref{tab:avg_results-comparsion} shows the coverage results of \IDFSP with both $\minF$ and $\maxF$ using using different heuristic functions ($\hMAX$, $\hFF$, and $\hADD$) against the other \FOND planners over both benchmark sets. Note that \IDFSP solved more tasks than the other planners. Namely, by comparing \IDFSP with \PRP and \MYND, note that \IDFSP with $\hADD$ outperforms \PRP and \MYND (in terms of solved tasks) with any of the three used heuristics.

Table~\ref{tab:results-comparsion} shows a detailed comparison between the best-evaluated variation of our algorithm against the best-evaluated variations of \PRP, \MYND, and \FONDSAT.
\IDFSP($\maxF$, $\hADD$) outperforms all the other \FOND planners in terms of solved tasks and planning time. Our best algorithm performed better than \PRP and \MYND over the NEW-\FOND benchmarks.
\FONDSAT also performed well for solving \FOND planning tasks over the NEW-\FOND benchmarks, as \citet{GeffnerG18_FONDSAT}~have shown. When comparing the \FOND planners in terms of policy size ($|\pi|$), on average, \FONDSAT is the planner that returns more compact policies.
However, we note that our algorithm and \MYND do not compact the policies using partial states, whereas \PRP and \FONDSAT do. Apart from some tasks for \doors, \faults, and \firstresp, \IDFSP($\maxF$, $\hADD$) has returned policies that are as compact as the ones returned by \PRP and \FONDSAT, see $|\pi|$ in Table~\ref{tab:results-comparsion}. 

Figures~3b,~3c, and~3d show a comparison among the \FOND planners with respect to planning time over all planning tasks for both benchmark sets. Planning tasks that \emph{timed out} are at the limit of x-axis and y-axis (300 seconds). Figure~3b shows that our algorithm is slower than \PRP for a substantial number of tasks, but \PRP \emph{timed out} for more tasks (most for the NEW-\FOND benchmark set shown as blue diamond). When comparing our algorithm with \MYND (Figure~3c), it is overall faster than \MYND and \emph{timed out} for fewer tasks. Figure~3d shows the planning time comparison between our algorithm and \FONDSAT. Our algorithm is faster and solves more tasks than \FONDSAT.

Figure~4 shows the number of solved tasks throughout the range of run-time for our algorithm (\IDFSP($\maxF$, $\hADD$)) against \PRP, \MYND, and \FONDSAT. When comparing the \FOND planners over all benchmark sets (Figure~4a), \IDFSP($\maxF$, $\hADD$) has more solved tasks than \MYND and \FONDSAT throughout all the range of run-time and is competitive with \PRP. Our algorithm (light-blue line) surpasses \PRP (red line) in terms of solved tasks after $\approx$ 200 seconds of planning time.
Over the NEW-\FOND benchmark set, Figure~4a shows that our algorithm \IDFSP($\maxF$, $\hADD$) outperforms all the other \FOND planners throughout all the range of run-time.

\begin{figure}[h!]
	\centering
	\begin{subfigure}[b]{1.03\empiricalfiguressize}
		\centering
		\includegraphics[width=\columnwidth]{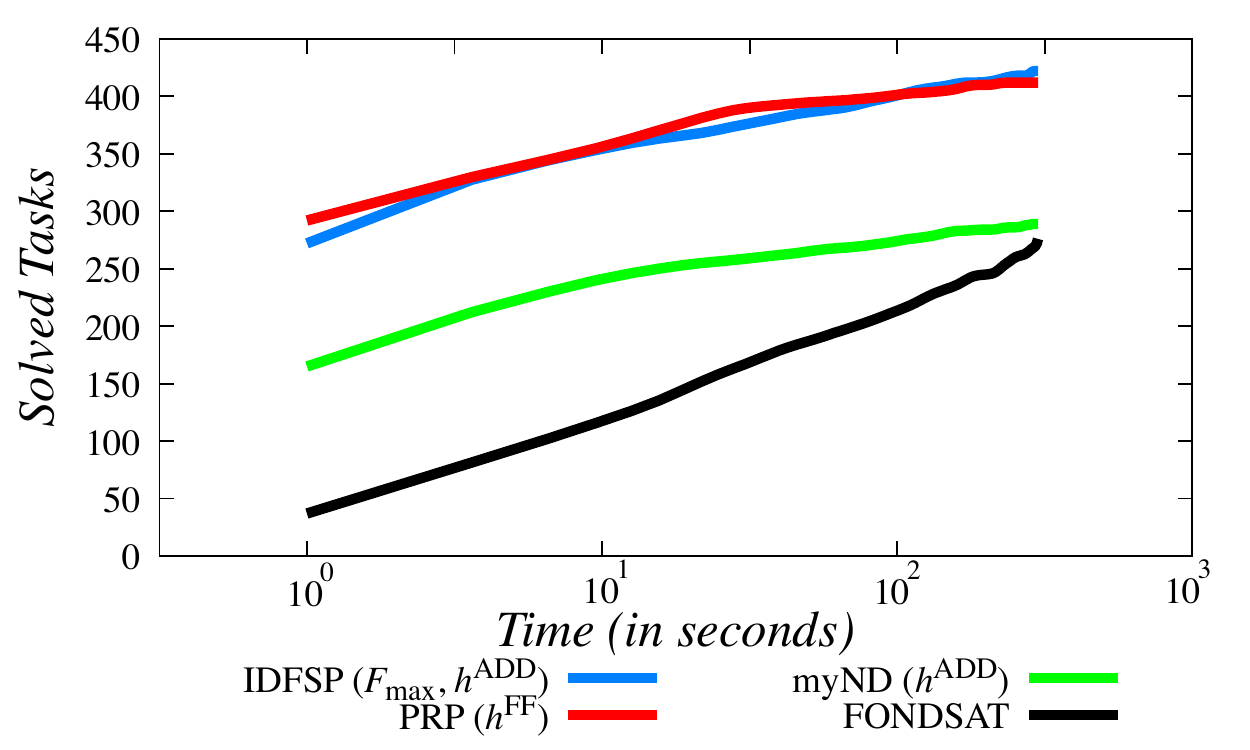}
		\caption{\scriptsize All benchmarks.}
		\label{fig:coverage_over_time_all}
	\end{subfigure}
	\begin{subfigure}[b]{1.03\empiricalfiguressize}
		\centering
		\includegraphics[width=\columnwidth]{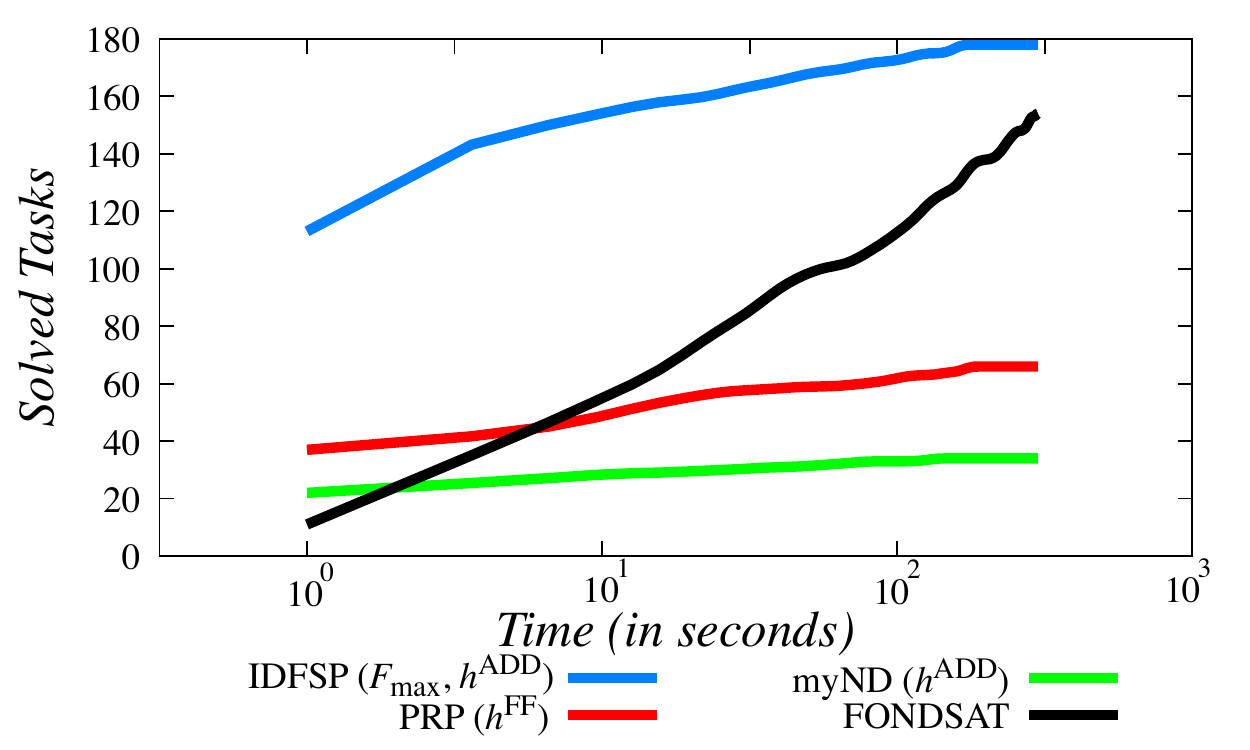}
		\caption{\scriptsize NEW-\FOND benchmarks.}
		\label{fig:coverage_over_time_interesting}
	\end{subfigure}
	\caption{Solved tasks throughout the range of run-time.}
	\label{fig:coverage_over_time}
\end{figure}

\section{Conclusions}\label{sec:conclusions}

We have developed a novel \emph{iterative depth-first search} algorithm that efficiently solves \FOND planning tasks. It considers more explicitly the non-determinism aspect of \FOND planning, and uses heuristic functions to guide the searching process. We empirically show that our algorithm can outperform existing planners concerning planning time and coverage.

As future work, we intend to investigate how to use the information gathered during previous iterations to make the following iterations of the searching more efficient. We also aim to investigate how to design more informed heuristic functions for \FOND planning. We aim to study the problem of designing algorithms to extract \emph{dual policy} solutions, when \emph{fairness} is not a valid assumption~\cite{Camanho_FONDknowpros16,GeffnerG18_FONDSAT,FONDPlus_RodriguezBSG21}. We also aim to investigate how to design domains and \FOND planning tasks that better capture the most significant characteristics of \FOND planning. These domains and tasks can be used to evaluate new planners.

\section*{Acknowledgments}

André acknowledges support from FAPERGS with projects 17/2551-0000867-7 and 21/2551-0000741-9, and Coordenação de Aperfeiçoamento de Pessoal de Nivel Superior (CAPES), Brazil, Finance Code 001.
Frederico acknowledges UFRGS, CNPq and FAPERGS for partially funding his research.
Ramon and Giuseppe acknowledge support from the ERC Advanced Grant WhiteMech (No. 834228) and the EU ICT-48 2020 project TAILOR (No. 952215). Giuseppe also acknowledges the JPMorgan AI Research Award 2021.


\bibliography{references}




\end{document}